\pdfoutput=1

\documentclass[11pt]{article}

\usepackage[]{ACL2023}

\usepackage{times}
\usepackage{latexsym}

\usepackage[T1]{fontenc}

\usepackage[utf8]{inputenc}

\usepackage{microtype}

\usepackage{inconsolata}

\usepackage{amsmath,bm}
\usepackage{inconsolata}
\usepackage{blindtext}
\usepackage{tcolorbox}
\usepackage{tabularx}

\usepackage{url}
\usepackage{multicol}
\usepackage{booktabs}
\usepackage{makecell}
\usepackage{amsmath, amssymb}
\usepackage{multirow}
\usepackage{mathtools}
\usepackage{enumitem}
\usepackage{float}
\usepackage{graphicx}
\usepackage{upgreek}
\usepackage{seqsplit}
\usepackage{color,soul}
\usepackage{arydshln}
\usepackage{placeins}
\usepackage{pifont}
\usepackage{amssymb}
\usepackage{bbm}
\usepackage{varwidth}

\usepackage{algorithm}
\usepackage{algpseudocode}
\usepackage{amsmath}
\usepackage{xcolor}

\usepackage{lipsum}  

\usepackage{amsmath,amsfonts,bm}

\def\eqref#1{equation~\ref{#1}}

\def\1{\bm{1}}

\DeclareMathAlphabet{\mathsfit}{\encodingdefault}{\sfdefault}{m}{sl}
\SetMathAlphabet{\mathsfit}{bold}{\encodingdefault}{\sfdefault}{bx}{n}

\DeclareMathOperator*{\argmax}{arg\,max}

\usepackage{caption}
\usepackage{subcaption}

\newcommand{\hlc}[2][yellow]{{\sethlcolor{#1}\hl{#2}}}

\title{Lossless Acceleration of Large Language Models with Hierarchical Drafting based on Temporal Locality in Speculative Decoding}

\author{Sukmin Cho$^1$
\quad Sangjin Choi$^1$
\quad Taeho Hwang$^2$
\quad Jeongyeon Seo$^2$
\quad Soyeong Jeong$^3$\\ 
\textbf{Huije Lee}$^2$
\quad \textbf{Hoyun Song}$^2$
\quad \textbf{Jong C. Park}$^2$
\quad \textbf{Youngjin Kwon}$^1$\thanks{\hspace{0.2cm}Corresponding author}\\ 
        School of Computing$^{1,2}$\quad Graduate School of AI$^3$ \\
        Korea Advanced Institute of Science and Technology\\ 
       \scriptsize{\texttt{\{smcho,sjchoi,yjkwon\}@casys.kaist.ac.kr}$^1$\quad\texttt{\{doubleyyh,yena.seo,starsuzi,huijelee,hysong,jongpark\}@kaist.ac.kr}$^{2,3}$}}

\begin{document}
\maketitle

\begin{abstract}
Accelerating inference in Large Language Models (LLMs) is critical for real-time interactions, as LLMs have been widely incorporated into real-world services.  Speculative decoding, a fully algorithmic solution, has gained attention for improving inference speed by drafting and verifying tokens, thereby generating multiple tokens in a single forward pass. However, current drafting strategies usually require significant fine-tuning or have inconsistent performance across tasks.
To address these challenges, we propose \textbf{Hierarchy Drafting} (HD)\footnote{\scriptsize \url{https://github.com/zomss/Hierarchy_Drafting}}, a novel lossless drafting approach that organizes various token sources into multiple databases in a hierarchical framework based on temporal locality. 
In the drafting step, HD sequentially accesses multiple databases to obtain draft tokens from the highest to the lowest locality, ensuring consistent acceleration across diverse tasks and minimizing drafting latency.
Our experiments on Spec-Bench using LLMs with 7B and 13B parameters demonstrate that HD outperforms existing lossless drafting methods, achieving robust inference speedups across model sizes, tasks, and temperatures.
\end{abstract}

\section{Introduction}
With the growing demand for accelerating Large Language Model (LLM) inference to enable efficient real-time human-LLM interactions, Speculative Decoding~\cite{BlockWise, SpecDecoding, SpecSampling} has gained attention for providing a fully algorithmic solution with minimal drawbacks.
While autoregressive decoding generates token by token, the decoding step in this method is divided into two substeps: \textit{drafting}, where likely tokens are sampled externally from a less complex model, and \textit{verifying}, where the sampled tokens are accepted or rejected by comparing with the LLM’s actual output.
By allowing the LLM to generate multiple accepted tokens in the verification phase, speculative decoding improves both the throughput and the latency of the LLM inference. 
Crucially, the efficiency of this approach depends on how draft tokens are generated, as performance gains hinge on the acceptance rate of these tokens~\cite{SpecSampling}.
Therefore, subsequent approaches to speculative decoding have focused on developing drafting strategies that sample tokens closely aligned with the target model.

\begin{figure}[t!]
\centering
\includegraphics[width=0.97\columnwidth]{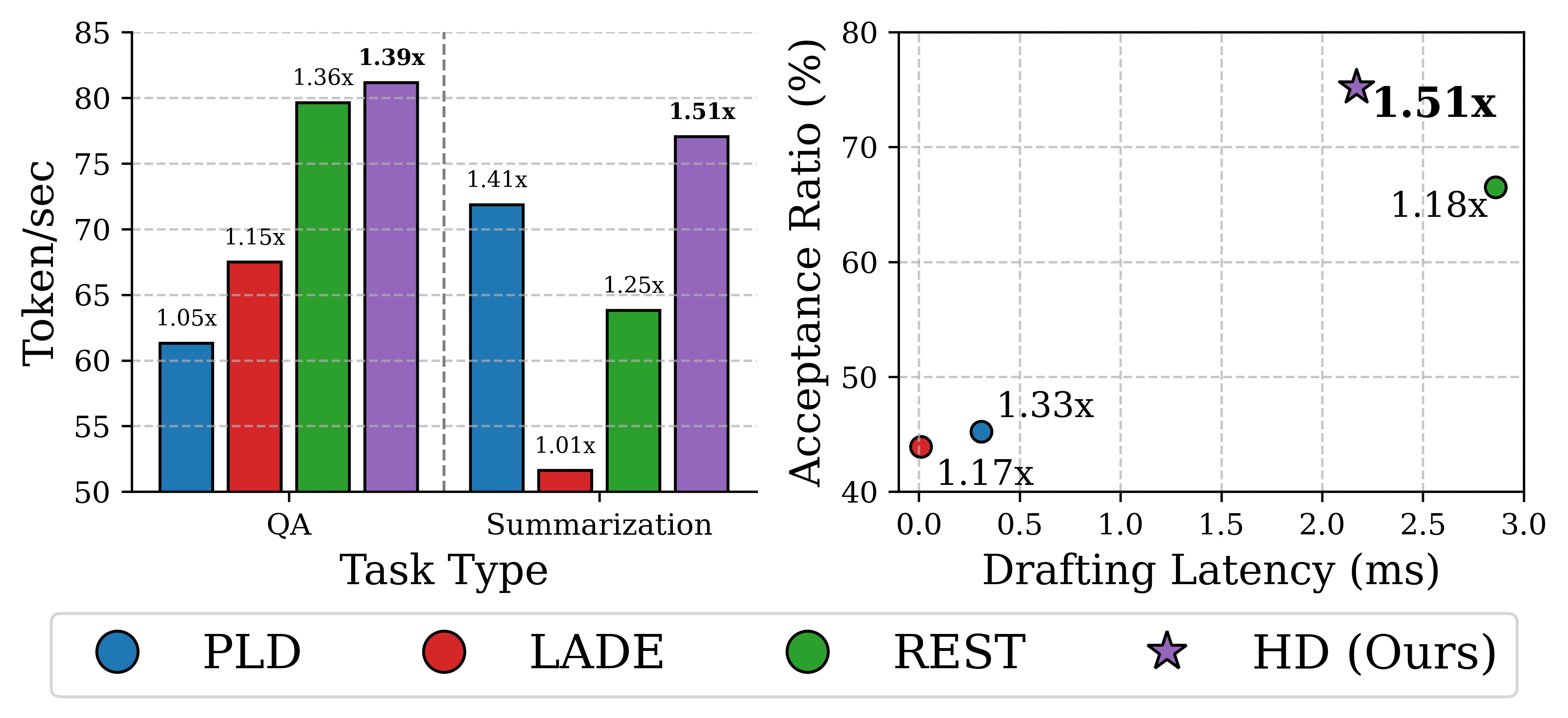}
\vspace{-.8em}
\caption{\small Result of database drafting methods on Spec-Bench~\cite{Spec_Survey} with Vicuna-7B~\cite{vicuna}. The values in the plot denote the speedup against autoregressive decoding. (Left) QA and summarization task performance. (Right) Acceptance ratio and drafting latency.}

\label{fig:motivation}
\vspace{-2.0em}
\end{figure}

Recent efforts in speculative decoding have focused on developing effective drafting methods, using LM-based approaches, such as using smaller models than LLM~\cite{DistilSpec, SpecInfer} or incorporating specialized branches within the LLM architecture~\cite{MEDUSA, EAGLE2}.
However, their applicability in real-world scenarios is limited by the significant overhead associated with fine-tuning for optimization.
First, smaller models for drafting must be fine-tuned, such as by distillation, to generate tokens similar to LLMs to achieve optimal performance regardless of the given tasks~\cite{DistilSpec, multilingual}.
In addition, current LLM families~\cite{Llama2, vicuna} do not offer models of an appropriate size for drafting, often necessitating training from scratch.
In branch-based drafting, which modifies its original LLM architecture, the computational cost for training such branches within LLM is significant due to gradient calculations across the entire model, even though most parameters remain frozen~\cite{MEDUSA, EAGLE2, EAGLE}.
For example, EAGLE~\cite{EAGLE}, one of the leading methods, needs 1-2 days of training on 2-4 billion tokens using 4 A100 GPUs to train the 70B model.

To address these limitations, this paper explores a lightweight, lossless drafting strategy: \textit{Database Drafting}, eliminating the need for parameter updates~\cite{PLD, LAD, REST}. 
Database drafting constructs databases from various token sources and fetches draft tokens from the database using previous tokens.
However, as previous work relies on a single database from a single source, the coverage of draft tokens is restricted, leading to inconsistent acceleration across different tasks, as depicted in the left side of Figure~\ref{fig:motivation}. 
For example, PLD~\cite{PLD}, which uses previous tokens as its source, shows strengths in the summarization, highly repeating the tokens in the earlier texts, yet it achieves only marginal speedups in QA, where fewer promising tokens are included in the prior text. 
A straightforward solution to improve coverage is incorporating diverse sources into a single database. 
However, increasing the database scale leads to higher drafting latency, resulting in additional overhead.
As shown in the right side of Figure~\ref{fig:motivation}, REST~\cite{REST}, which uses the largest database, accurately predicts future tokens but suffers from significant latency, negating its high acceptance ratio benefits. 
Therefore, this paper proposes a solution to these limitations: \textit{Utilize diverse token sources simultaneously for robust performance and minimal overhead.}

With this objective in mind, we propose a simple yet effective solution: \textbf{Hierarchy Drafting} (HD), which integrates diverse token sources into a hierarchical framework. 
Our proposed method is inspired by the memory hierarchy system, which prioritizes data with high \textit{temporal locality} in the memory access for performance optimization~\cite{hierarchy}.
Therefore, HD groups draft tokens from diverse sources based on their temporal locality---the tendency for some tokens to reappear within or across generation processes. 
For example, when an LLM solves a math problem like, ‘\textit{The vertices of a triangle are at points (0, 0), (-1, 1), and (3, 3). What is the area of the triangle?}’, the coordinates frequently repeat within only a generation process for a given query but not across other generation processes.
In a related sense, phrases commonly generated by LLMs, such as ‘\textit{as an AI assistant}’, or frequent grammatical patterns exhibit relatively moderate locality, often appearing across different generation processes. 

Based on their temporal locality, the multiple databases of HD organize them into \textit{context-dependent database}, which stores tokens with high temporal locality for a given context; \textit{model-dependent database}, which captures frequently repeated phrases by LLMs across generations; and \textit{statistics-dependent database}, which contains statistically common phrases with slightly lower locality across processes than those in the model-dependent database.
During inference, HD accesses the databases in order of temporal locality, prioritizing tokens with high locality by starting with context-dependent, then model-dependent, and finally statistics-dependent databases until a sufficient number of draft tokens are obtained to convey to the LLM for verification.

This strategy has two benefits: firstly, increasing drafting accuracy by leveraging temporal locality and
secondly, reducing the overhead from drafting latency, as the scale of the databases is inversely correlated with the degree of locality—tokens with high locality are rarer. Thus, starting with the smaller context-dependent database for drafting tokens is more accurate and faster than using the larger statistics-dependent database alone.
Also, our hierarchical framework can encompass other database drafting methods owing to its \textit{plug-and-play} nature, making it easy to integrate diverse drafting sources based on their temporal locality.

We evaluate HD and other database drafting methods using widely adopted LLMs, Llama-2~\cite{Llama2} and Vicuna~\cite{vicuna}, on Spec-Bench~\cite{Spec_Survey}, a benchmark designed to assess effectiveness across diverse tasks.
Our proposed method, HD, outperforms other methods in our experiment and consistently achieves significant inference speedup across various settings, including model size, temperature, and tasks.
We also analyze how the hierarchical framework adaptively selects the appropriate database for each task while minimizing draft latency, aligning with our design goals.

Our contributions in this paper are threefold:
\vspace{-0.1in}
\begin{itemize}[itemsep=0.3mm, parsep=1pt, leftmargin=*]
    \item We identify the limitations of existing speculative decoding methods, which require additional fine-tuning or deliver inconsistent acceleration gains.
    \item We introduce a novel database drafting method, Hierarchy Drafting (HD), incorporating diverse token sources into the hierarchical framework for robust performance with minimizing overhead.
    \item We demonstrate that HD consistently achieves significant acceleration gains across various scenarios compared to other lossless methods.
\end{itemize}

\section{Related Work}
We now introduce speculative decoding and lossless drafting strategies based on the database.

\paragraph{Speculative Decoding} 
Speculative decoding is a novel approach that accelerates LLM inference by minimizing the number of forward passes required, thereby reducing total latency~\cite{BlockWise, SpecDecoding, SpecSampling}. The core concept is that tokens, such as frequent phrases, can be predicted with high confidence using simpler models, enabling the generation of multiple tokens at once. \citet{BlockWise} introduced the \textit{Draft-then-Verify} paradigm, dividing each decoding step into two sub-steps: drafting multiple tokens from draft models and verifying them against LLM outputs in parallel. This concept has been expanded to accurately speculate the future tokens along with supporting sampling strategy~\cite{seq2seq, SpecDecoding, SpecSampling}. 

\paragraph{Types of Drafting Method}
The straightforward approach for the drafting strategy of speculative decoding involves using an additional language model (LM) specialized for drafting~\cite{SpecDecoding, SpecSampling, SpecInfer, DistilSpec}. 
To ensure effective drafting, such LMs must follow the target model's generation pattern and be smaller to minimize additional latency costs.
LMs with parameter sizes under a billion are typically preferred for drafting, but currently, widely used LLM families do not usually have appropriate models. 
For example, the smallest officially available Llama-2 model~\cite{Llama2}, with 7 billion parameters, is too large and inefficient for drafting purposes.
Therefore, such methodologies often require training overhead to get the suitable LM for the targeted LLM, such as the distilled models from the target models~\cite{SpecInfer} or lightweight models trained for mobile devices~\cite{TinyLlama}.

Instead of using a separate language model for drafting, some approaches enhance the drafting capabilities of the target model itself~\cite{MEDUSA, EAGLE2, EAGLE, Hydra}. 
In this line of work, the additional layer or branch in the target model is integrated into the target model to predict several subsequent tokens more than the very next token based on the last hidden states of given inputs.
Following \citet{BlockWise}, which exploits multiple heads for parallel decoding, Medusa~\cite{MEDUSA} first integrates additional decoding heads into the target model.
Subsequently, the branch-based drafting methodologies~\cite{EAGLE2, EAGLE, Hydra} show remarkable effectiveness in sampling appropriate future tokens with achieving state-of-the-art results.
However, integrating these layers or branches still requires significant training overhead. 
To sum up, branch-based drafting methods achieve remarkable speedup gains yet require additional computational costs, which are not trivial and are a new type of overhead for implementing speculative decoding.

\paragraph{Database Drafting}  
Database drafting eliminates training costs by retrieving draft tokens for previous inputs from a database rather than relying on smaller LMs or additional architectural branches. 
The database stores token pairs, with prefix tokens as keys and subsequent tokens as values.
The sources of these databases vary across different methods, with each method relying on its own unique database source. Some approaches utilize input prompt tokens as draft sources, which is particularly effective for tasks like summarization or retrieval-augmented generation, where input tokens are frequently repeated during generation~\cite{PLD, InfwRef}. Another method retrieves draft tokens from large text corpora by leveraging language patterns~\cite{REST}. 
Although retrieval from large corpora introduces some latency overhead, the acceleration gained from accurate drafting typically outweighs this, resulting in faster inference overall.
Additionally, LLMs can serve as sources for database drafting by generating tokens stored in the database, either through parallel decoding~\cite{ParallelDecoding, LAD} or token recycling~\cite{trashintotreasure}, where tokens are relevant to the current generation process. 
Finally, the previously generated texts by LLMs can be served as draft token sources because LLMs frequently reuse specific phrases or words~\cite{StagedSpec}.
Each source offers distinct strengths in predicting future tokens in certain scenarios, yet these strengths can become weaknesses in others. Therefore, it is worth noting that reliance on a single source may lead to limitations.

\begin{table}[t]
    \caption{\small Details of current database drafting methods with Vicuna-7B on Spec-Bench. Database scale measures the number of draft token sequences in each database. Drafting latency measures the average latency for the drafting step.} \label{tab:related_work}
    \vspace{-0.7em}
    \centering
    \renewcommand{\arraystretch}{1}
\resizebox{.99\columnwidth}{!}{
    \begin{tabular}{l  c c c}
    \toprule
       \textbf{Method} & \textbf{Database Scale} & \textbf{Drafting Latency}  & \textbf{\#Token/Sec} \\ \midrule
       \textbf{PLD}~\cite{PLD} & 519 \(\pm\) 423 & 0.31 \(\pm\) 0.06 ms& 74.51 \(\pm\) 23.09 \\
       \textbf{LADE}~\cite{LAD} & 354 \(\pm\) 258 & < 0.01 ms & 66.34 \(\pm\) 14.09\\ 
       \textbf{REST}~\cite{REST} & 200M & 2.86 \(\pm\) 6.43 ms & 66.04 \(\pm\) 14.72\\ \bottomrule
    \end{tabular}
} 
\vspace{-1.0em}
\end{table}

Table~\ref{tab:related_work} shows the experimental results of current database drafting methods, which construct their databases from a single source. 
Specifically, PLD~\cite{PLD} exhibits the highest speedup compared to other approaches but also shows a significant standard deviation in speedup gains. This variability is attributed to the limited and uneven sizes of the databases, leading to inconsistent acceleration across the generation process.
In contrast, LADE~\cite{LAD} achieves an impressively low drafting latency—less than 0.01 ms. However, this remarkable value does not translate to significant acceleration due to its small database size, akin to PLD. 
However, increasing database size alone, as demonstrated by REST~\cite{REST}, does not provide a viable solution for improving the effectiveness of database drafting. While a larger database scale can improve the accuracy of the drafting step, it also leads to higher latency since retrieving tokens from a larger database introduces additional processing overhead.

Therefore, to address the limitations of current lossless drafting methods relying on a single source, we propose integrating diverse sources into a hierarchical framework, aiming to harness each source's strengths more effectively with minimal overhead.

\begin{figure*}[t!]
\centering
\includegraphics[width=\textwidth]{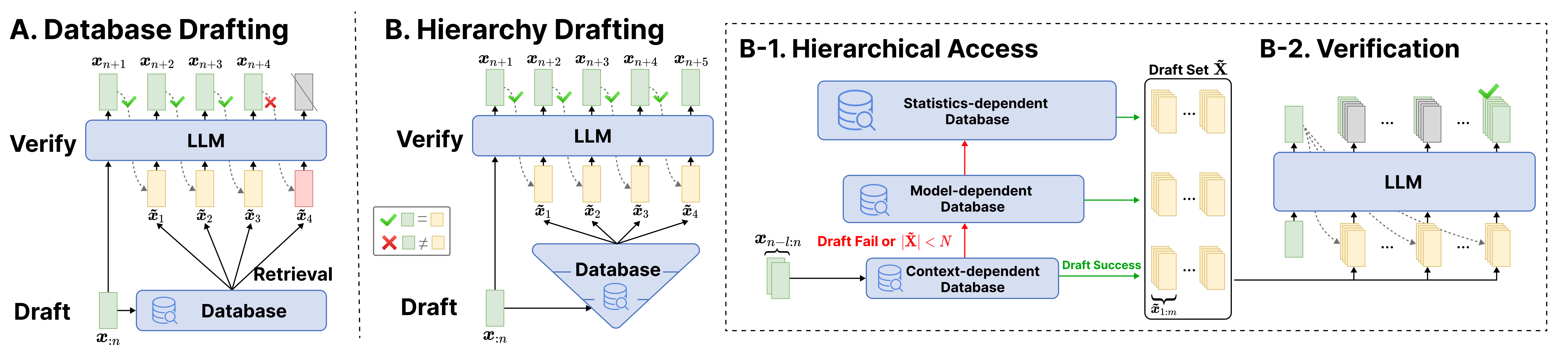}
\vspace{-.8em}
\caption{\small Overview of database drafting and our proposed method, Hierarchy Drafting (HD). \textit{A.} Previous database drafting methods retrieve draft tokens from a single database constructed from a single source, leading to inconsistent acceleration gains across different scenarios. \textit{B.} HD, however, leverages multiple databases encompassing diverse sources to improve token coverage, ensuring consistent performance.
\textit{B-1.} During the drafting process, databases are accessed sequentially from the smallest to the largest, based on the temporal locality of the token sequences.
\textit{B-2.} Multiple draft token sequences are verified in parallel, and the sequence with the highest number of accepted tokens is finally selected as the generated output.}
\label{fig:overview}
\vspace{-1.3em}
\end{figure*}


\section{Method}

We begin by formally defining speculative decoding and database drafting and present our proposed method, Hierarchy Drafting (HD), which addresses the limitations of database drafting methods.

\subsection{Preliminary}

\paragraph{Speculative Decoding} 
At each step of speculative decoding, multiple tokens \(\tilde{\bm{x}}_{1:m}\) (i.e., draft token sequence) are drafted from an approximate model \(\mathcal{M}_q\) to predict future tokens of LLM \(\mathcal{M}_p\) (i.e., target model) for previous text tokens \(\bm{x}_{\leq t}\):
\begin{align}
    \tilde{\bm{x}}_{1:m} &\sim_m \mathcal{M}_q(\bm{x}_{\leq t}).
\end{align}

All draft token sequence \(\tilde{\bm{x}}_{1:m}\) are verified against the actual output of \(\mathcal{M}_p\). For example, in the greedy decoding, the tokens \(\bm{x}'_{t+1:t+m}\) are obtained for a given \(\tilde{\bm{x}}_{1:m}\) and \(\bm{x}_{\leq t}\) by solving the following equations in parallel:
\begin{align}
\begin{cases}
    x'_{t+1} &= \argmax P_{\mathcal{M}_p}(x | \bm{x}_{\leq t}), \\
    x'_{t+2} &= \argmax P_{\mathcal{M}_p}(x | \tilde{x}_{1}, \bm{x}_{\leq t}),\\
    &\dots \\
    x'_{t+m} &= \argmax P_{\mathcal{M}_p}(x | \tilde{\bm{x}}_{1:m}, \bm{x}_{\leq t}).
\end{cases}
\end{align}
Each token \(x'_{t+i}\) is verified against the corresponding draft token \(\tilde{x}_{t+i}\), starting from \(i = 0\) until the verification fails or \(i = m\) is reached.
To enhance the likelihood of acceptance, multiple draft token sequences \(\bm{\tilde{X}} = \{\tilde{\bm{x}}^i\}_{i=1}^N\) (i.e., draft set) are verified in parallel.
The specialized attention mask implements the parallel verification of the draft set, not causal attention mask~\cite{LAD, SpecInfer}.
In the sampling strategy, speculative sampling~\cite{SpecSampling} is commonly used to accept more tokens while maintaining identical output distributions of the target model.
In summary, the generation step is divided into two sub-steps with a single forward pass of the target model. The multiple accepted tokens are generated simultaneously, compressing the overall decoding process.

\paragraph{Database Drafting}
As shown on the left side of Figure~\ref{fig:overview}, the methods included in database drafting exploit the database \(\mathcal{D}\), having the prefix tokens as the key and the subsequent tokens as the value. Per each step of the generation process, the draft token sequence \(\bm{\tilde{x}}_{1:m}\) is retrieved from database \(\mathcal{D}\) for given previous tokens \(\bm{x}_{t-l:t}\):
\begin{align}
    \bm{\tilde{x}}_{1:m} \in \bm{\tilde{X}} &= \texttt{Ret}(\bm{x}_{t-l:t};\mathcal{D}),
\end{align}
where \(l\) and \(m\) are the length of previous tokens and draft token sequence. Subsequently, the verifying step is the same as other methods.

\subsection{Hierarchy Drafting}

We introduce Hierarchy Drafting (HD), which organizes tokens from diverse sources into three databases based on temporal locality and accesses them in order from the smallest to the largest scale. The overview and decoding process are depicted on the right side of Figure~\ref{fig:overview}.

\begin{figure}[t!]
\centering
\includegraphics[width=0.97\columnwidth]{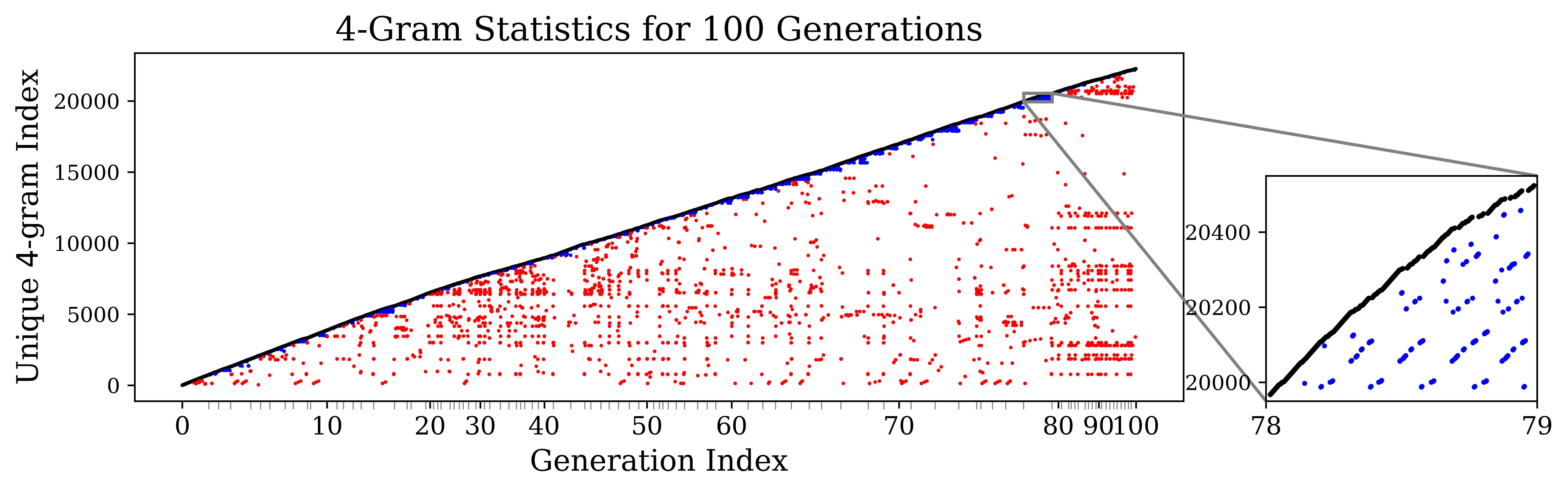}
\includegraphics[width=0.97\columnwidth]{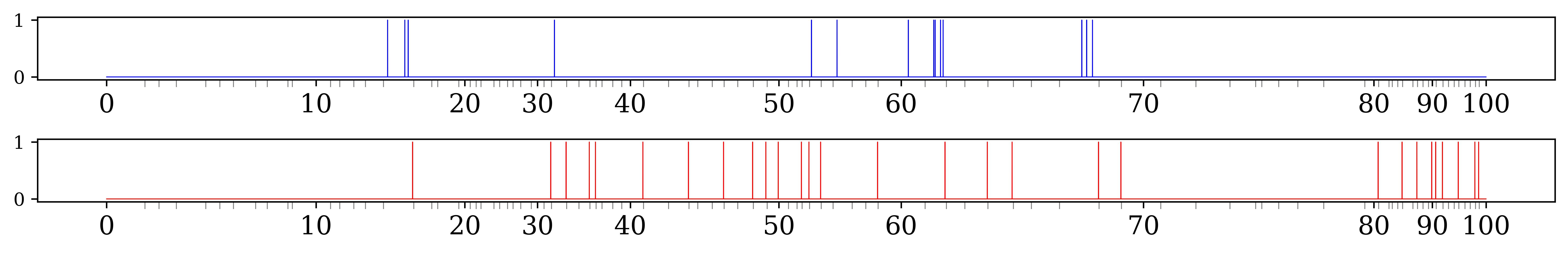}
\vspace{-.8em}
\caption{\small (Upper) 4-gram statistics for 100 generations of Llama-2-7b. The x-axis shows the order of 4-grams across 100 generations, with major ticks marking generation steps. The y-axis represents unique 4-gram indices. Red dots indicate 4-grams from previous processes, while blue dots represent those from the current process. (Lower) Frequency analysis of two 4-grams, represented by red and blue dots, respectively.}
\label{fig:generation}
\vspace{-1.5em}
\end{figure}

\paragraph{Observation: Temporal Locality}
The main idea behind database drafting is that some tokens are easy to retrieve from the database because they exhibit temporal locality—meaning they tend to be repeated within or across the generation processes. 
However, note that not all draft token sequences share the same level of temporal locality during generation. 
We analyze the pattern of unique 4-grams during 100 text generations on Spec-Bench~\cite{Spec_Survey}, as shown in Figure~\ref{fig:generation}. The results reveal that certain 4-grams are frequently repeated and exhibit varying locality levels.
Specifically, the blue dots and the right small plot in Figure~\ref{fig:generation} illustrate local redundancy, where the same 4-gram appears multiple times within a single generation step. This reflects high temporal locality within a single generation rather than across multiple generations. In contrast, the red dots in Figure~\ref{fig:generation} highlight a pattern where the model repeatedly generates the same 4-grams at different stages of the generation process, illustrating its tendency to reuse familiar sequences over time.
Additionally, the lower plot of Figure~\ref{fig:generation} presents the frequency study of sampled red and blue dots, demonstrating that some tokens exhibit high temporal locality within a specific context, while others maintain consistent locality across generation processes.
Therefore, given the varying temporal locality of tokens throughout the generation process, drafting steps should prioritize tokens with higher temporal locality over others.


\paragraph{Database Design}  
 Based on the temporal locality of draft token candidates, we design three types of databases to categorize them. 
\textbf{1) Context-dependent DB} (\(\mathcal{D}_c\)) contains tokens highly relevant to the specific context of the generation process, such as the blue dots in the Figure~\ref{fig:generation}. 
This includes tokens from the input prompt, tokens generated through parallel decoding, tokens discarded during the generation process, and others that are highly relevant to a given context.  
\(\mathcal{D}_c\) is lookup table with the prefix tokens, \(\bm{x}_{1:l}\), as the key and the subsequent tokens, \(\bm{x}_{l:l+m}\), as the value.
Also, \(\mathcal{D}_c\) is consistently updated during each forward step and initialized when the following generation process is started. 
The database follows the Least Recently Used (LRU) policy for draft sequence updates.
\textbf{2) Model-dependent DB} (\(\mathcal{D}_m\)) stores tokens frequently generated by LLM regardless of context, as represented by the red dots in Figure~\ref{fig:generation}.
Top-$k$ frequently generated token sequences, $\bm{x}_{1:l+m}$, are sampled from the model-generated texts, with $\bm{x}_{1:l}$ as the key and $\bm{x}_{l+1:l+m}$ as the value.
For \(\mathcal{D}_c\) and \(\mathcal{D}_m\), the maximum size of values for a single key is the same as the maximum draft set size \(N\). 
\textbf{3) Statistics-dependent DB} (\(\mathcal{D}_s\)) draws its tokens from large text corpora to capture universal phrases commonly used in the language. 
Although these tokens are frequent, they occur less consistently across processes than those in \(\mathcal{D}_m\).
To efficiently retrieve the sequence from a large corpus, we utilize a suffix array~\cite{suffix_array} following the implementation of~\citet{REST}.
Implementation details are in \S\ref{sec:experiement}.

Our database design yields three distinct advantages. First, it integrates diverse sources into multiple databases, enabling us to leverage each source’s strengths for robust acceleration across various tasks. Then, each database’s size decreases as the tokens’ temporal locality increases since tokens with higher locality are rarer, providing an opportunity to optimize drafting latency. Finally, the design is \textit{plug-and-play}, easily integrating additional token sources by assigning them to the appropriate database based on their temporal locality.

\setlength{\textfloatsep}{1em}

\begin{algorithm}[t!]
\caption{\small Decoding Process with Hierarchy Drafting}\label{alg:HD_process}
\small
\begin{algorithmic}[1]
\Require Target LLM $\mathcal{M}_p$, databases $(\mathcal{D}_c, \mathcal{D}_m, \mathcal{D}_s)$, input text sequence $\bm{x}_{\le t}$, target sequence length $T$, the size of prefix tokens $l$, the size of draft token sequence $m$, the size of draft set $N$;
\State $n \leftarrow t$\;
\While{$n < T$ and \texttt{[EOS]} $ \notin \bm{x}_{1:n}$}
    \State \textcolor{blue}{// \textit{Drafting Step: Hierarchical access to three databases until the size of the draft set $\bm{\tilde{X}}$ is $N$.}}
    \State $\bm{\tilde{X}} \leftarrow \texttt{Ret}(\bm{x}_{n-l:n};\mathcal{D}_c)$
    \If{$|\bm{\tilde{X}}| < N$}
        \State $\bm{\tilde{X}} \leftarrow \bm{\tilde{X}} \cup \texttt{Ret}(\bm{x}_{n-l:n};\mathcal{D}_m)$
    \EndIf 
    \If{$|\bm{\tilde{X}}| < N$}
        \State $\bm{\tilde{X}} \leftarrow \bm{\tilde{X}} \cup \texttt{Ret}(\bm{x}_{n-l:n};\mathcal{D}_s)$
    \EndIf 
    \State \textcolor{blue}{// \textit{Verification Step: Verify the draft token sequence in $\bm{\tilde{X}}$ and generate additional tokens for updating $\mathcal{D}_c$.}}
    \State $\bm{x}_{n:n+i}, \bm{\hat{x}} \sim_i \mathcal{M}_p(\bm{x}_{\le n}, \bm{\tilde{X}})$
    \State $\mathcal{D}_c \leftarrow \text{Update}(\mathcal{D}_c, \bm{\hat{x}})$
    \State $n \gets n+i$
\EndWhile
\end{algorithmic}
\end{algorithm}

\paragraph{Hierarchical Access}
Using the three databases designed with the temporal locality in mind, we retrieve draft token sequence \(\bm{\tilde{x}}_{1:m}\) for the given previous input \(\bm{x}_{t-l:t}\).
Database access order is based on the degree of temporal locality within the current generation process; thereby, the access starts with \(\mathcal{D}_c\).
Access then proceeds to \(\mathcal{D}_m\), which has high locality across generations, and finally \(\mathcal{D}_s\), with moderate locality across generations, until draft set \(\bm{\tilde{X}}\) accumulates a sufficient number of candidates as pre-defined hyperparameter \(N\).
These accesses leverage the locality of the draft token sequence to enhance drafting accuracy and minimize latency overhead, preserving the benefits of drafting.

\paragraph{Decoding Process}
We introduce the inference process of speculative decoding with our proposed method, HD. 
First, for a given previous input \(\bm{x}_{t-l, t}\), we acquire the set of draft token \(\bm{\tilde{X}}\) from the three databases with hierarchical access. 
Then, the target LLM \(\mathcal{M}_p\) verifies the draft token sequences while simultaneously generating the additional tokens \(\bm{\hat{x}}\). 
These tokens are used to update the context-dependent DB either through parallel decoding~\cite{ParallelDecoding, LAD} or by recycling wasted tokens~\cite{trashintotreasure}. 
These processes are repeated iteratively until either the \texttt{[EOS]} token is generated or the sequence reaches the pre-defined maximum length \(T\).
Details of the decoding are depicted in Algorithm~\ref{alg:HD_process}.

\section{Experimental Setup}
\label{sec:experiement}
We introduce the details of the experiment setups conducted to evaluate the effectiveness of HD.

\paragraph{Dataset} 
We exploit Spec-Bench~\cite{Spec_Survey}, a comprehensive benchmark to evaluate speculative decoding across various tasks. Specifically. the collected datasets are MT-bench~\cite{vicuna} for Multi-turn Conversation, WMT14 DE-EN~\cite{translation} for Translation, CNN/Daily Mail~\cite{summarization} for Summarization, Natural Question~\cite{QA} for Question Answering, GSM8K~\cite{math} for Math Reasoning, DPR~\cite{RAG} for RAG. Each task has 80 instances, making a total of 480 generations.

\paragraph{Model}
We utilize two LLM families: \textbf{Vicuna-v1.3-\{7,13,33\}B}~\cite{vicuna} and \textbf{Llama-2-chat-\{7,13\}B}~\cite{Llama2} to demonstrate the effectiveness of the proposed method. 

\begin{table*}[t]
\caption{\small Results of Hierarchy Drafting and various database drafting methods on Spec-Bench. The best results are \textbf{bold}.}
\vspace*{-1mm}
\centering
    \small
    \renewcommand{\arraystretch}{1}
\resizebox{\textwidth}{!}{
        \begin{tabular}{ll cccc cccc cccc cccc cccc
        }
        \toprule
        & & \multicolumn{4}{c}{\textbf{Vicuna-7B-v1.3}} & \multicolumn{4}{c}{\textbf{Vicuna-13B-v1.3}} & \multicolumn{4}{c}{\textbf{Vicuna-33B-v1.3}} & \multicolumn{4}{c}{\textbf{Llama-2-7B-chat}} & \multicolumn{4}{c}{\textbf{Llama-2-13B-chat}}\\ 
        \cmidrule(l{2pt}r{2pt}){3-6}
        \cmidrule(l{2pt}r{2pt}){7-10}
        \cmidrule(l{2pt}r{2pt}){11-14}
        \cmidrule(l{2pt}r{2pt}){15-18}
        \cmidrule(l{2pt}r{2pt}){19-22}
         & \textbf{Method} & D.L (ms) & \(\alpha\) (\%) & \(\tau\) & Speedup & D.L (ms) & \(\alpha\) (\%) & \(\tau\)  & Speedup & D.L (ms) & \(\alpha\) (\%) & \(\tau\) & Speedup & D.L (ms) & \(\alpha\) (\%) & \(\tau\) & Speedup & D.L (ms) & \(\alpha\) (\%) & \(\tau\) & Speedup \\ \midrule
        \multirowcell{5.3}[-0.0ex][c]{\rotatebox[origin=c]{90}{\(T=0.0\)}} & \textbf{AR} & - & - & 1.00 & 1.00x & - & - & 1.00 & 1.00x & - & - & 1.00 & 1.00x & - & - & 1.00 & 1.00x & - & - & 1.00 & 1.00x \\
        & \textbf{PLD} & 0.31 & 45.22 & 1.62 & 1.32x & 0.31 & 44.49 & 1.57 & 1.36x & 0.49 & 40.55 & 1.47 & 1.31x & 0.31 & 35.02 & 1.36 & 1.18x & 0.30 & 35.63 & 1.36 & 1.18x \\
        & \textbf{LADE} & \textbf{<0.01} & 43.92 & 1.64 & 1.18x & \textbf{<0.01} & 43.59 & 1.63 & 1.21x & \textbf{<0.01} & 45.34 & 1.61 & 1.24x &  \textbf{<0.01} & 50.03 & 1.59 & 1.22x & \textbf{<0.01} & 49.96 & 1.58 & 1.25x\\
        & \textbf{REST} & 2.85 & 66.51 & 1.82 &  1.17x & 3.12 & 66.80 & 1.82 & 1.38x & 3.18 & 66.54 & 1.80 & 1.33x &  2.85 & 72.33 & 1.92 & 1.38x & 3.16 & 72.48 & 1.92 & 1.52x \\
        \noalign{\vskip 0.3ex}\cdashline{2-22}\noalign{\vskip 0.7ex}
        & \textbf{HD} (Ours) & 2.17 & \textbf{75.21} & \textbf{2.38} & \textbf{1.51x} & 2.30 & \textbf{75.05} & \textbf{2.30} & \textbf{1.58x} & 1.66 & \textbf{73.56} & \textbf{2.19} & \textbf{1.63x} & 2.18 & \textbf{79.72} & \textbf{2.31} & \textbf{1.64x} & 2.33 & \textbf{79.63} & \textbf{2.29} & \textbf{1.70x} \\ \midrule
        \multirowcell{4.3}[-0.0ex][c]{\rotatebox[origin=c]{90}{\(T=1.0\)}} & \textbf{AR} & - & - & 1.00 & 1.00x & - & - & 1.00 & 1.00x & - & - & 1.00 & 1.00x & - & - & 1.00 & 1.00x & - & - & 1.00 & 1.00x \\
        & \textbf{LADE} & \textbf{0.01} & 38.09 & 1.49 & 1.11x & \textbf{0.01} & 40.27 & 1.51 & 1.16x & \textbf{0.02} & 44.10 & 1.57 & 1.26x &  \textbf{0.01} & 48.71 & 1.55 & 1.20x & \textbf{0.01} & 49.37 & 1.56 & 1.23x \\
        & \textbf{REST} & 2.79 & 66.91 & 1.84 & 1.17x & 3.09 & 67.61 & 1.84 & 1.39x & 3.22 & 66.98 & 1.81 & 1.38x &  2.81 & 73.00 & 1.93 & 1.39x & 3.02 & 72.85 & 1.93 & 1.53x\\
        \noalign{\vskip 0.3ex}\cdashline{2-22}\noalign{\vskip 0.7ex}
        & \textbf{HD} (Ours) & 2.41 & \textbf{69.11} & \textbf{2.06} & \textbf{1.32x} & 2.40 & \textbf{70.61} & \textbf{2.11} & \textbf{1.44x} & 1.72 & \textbf{71.00} & \textbf{2.09} & \textbf{1.58x} & 2.15 & \textbf{78.88} & \textbf{2.25} & \textbf{1.54x} & 2.30 & \textbf{79.40} & 2.29 & \textbf{1.63x}\\ \bottomrule
        
        \end{tabular}
}
\label{tab:main}
\vspace{-1.3em}
\end{table*}

\paragraph{Baseline Method}
We compare our proposed method, \textbf{HD}, with autoregressive decoding and various database drafting methods to validate its effectiveness. Specifically, \textbf{1) Autoregressive decoding (AR)} serves as an indicator for measuring acceleration gains. We also include \textbf{2) PLD}\footnote{PLD is included only in the greedy setting due to its official repository's lack of temperature sampling support.}~\cite{PLD}, utilizing previous input prompts as a database, \textbf{3) LADE}~\cite{LAD}, employing parallel decoding via a Jacobian iteration method, and \textbf{4) REST}~\cite{REST}, which retrieves draft tokens from a large text corpus. 


\paragraph{Evaluation Metric} 
We utilize a variety of metrics to evaluate drafting overhead, drafting accuracy, and acceleration gain. To measure drafting overhead, we use \textbf{1) Drafting Latency}, which refers to the time taken to fetch draft tokens. Following~\citet{DistilSpec}, the drafting accuracy is assessed using \textbf{2) Acceptance Ratio (\(\alpha\))} and \textbf{3) Mean Accepted Tokens (\(\tau\))}. The acceptance ratio (\(\alpha\)) represents the ratio of accepted tokens to total tokens, while the mean accepted tokens (\(\tau\)) denotes the expected number of accepted tokens per step. Finally, acceleration gain is measured using the \textbf{4) Speedup Ratio}, which compares \#tokens/sec of each method from autoregressive decoding.

\paragraph{Implementation Detail}
The proposed method, HD, is configured with the hyperparameters \(l\), \(m\), \(N\), and \(T\) set to 2, 4, 7, and 1024, respectively. 
Specifically, \(l\) denotes the length of the previous tokens used as the database key, and \(m\) represents the length of the draft sequence used as the database value. Finally, \(N\) specifies the size of the draft set passed to the LLM for verification. 
To adopt a sampling strategy, we exploit speculative sampling ~\cite{SpecSampling} by setting draft probability as 1.0.
For the context-dependent database (\(\mathcal{D}_c\)), the previous input tokens and the tokens generated via parallel decoding are included. 
For parallel decoding, we follow the implementation proposed by LADE~\cite{LAD}, which allows simultaneous processing of the parallel decoding and verification branches.
Therefore, following the implementation of LADE, the verifying step is based on \textit{n}-gram.
For the model-dependent database (\(\mathcal{D}_m\)), we collect LLM-generated texts from the English portion of the OASST training set~\cite{OASST}, using a 7B model from the targeted LLM family.
A total of 39,283 texts were generated, from which we sampled the 100k most frequent token sequences.
Lastly, for the statistics-dependent database (\(\mathcal{D}_s\)), we adopt the setting of REST~\cite{REST}, utilizing data sourced from UltraChat~\cite{UltraChat}, with the database size being approximately 12GB.
More details are in Appendix~\ref{app:impl}.

\begin{figure*}[t!]
\centering
\begin{minipage}{0.3\textwidth}
    \centering
    \includegraphics[width=0.99\textwidth]{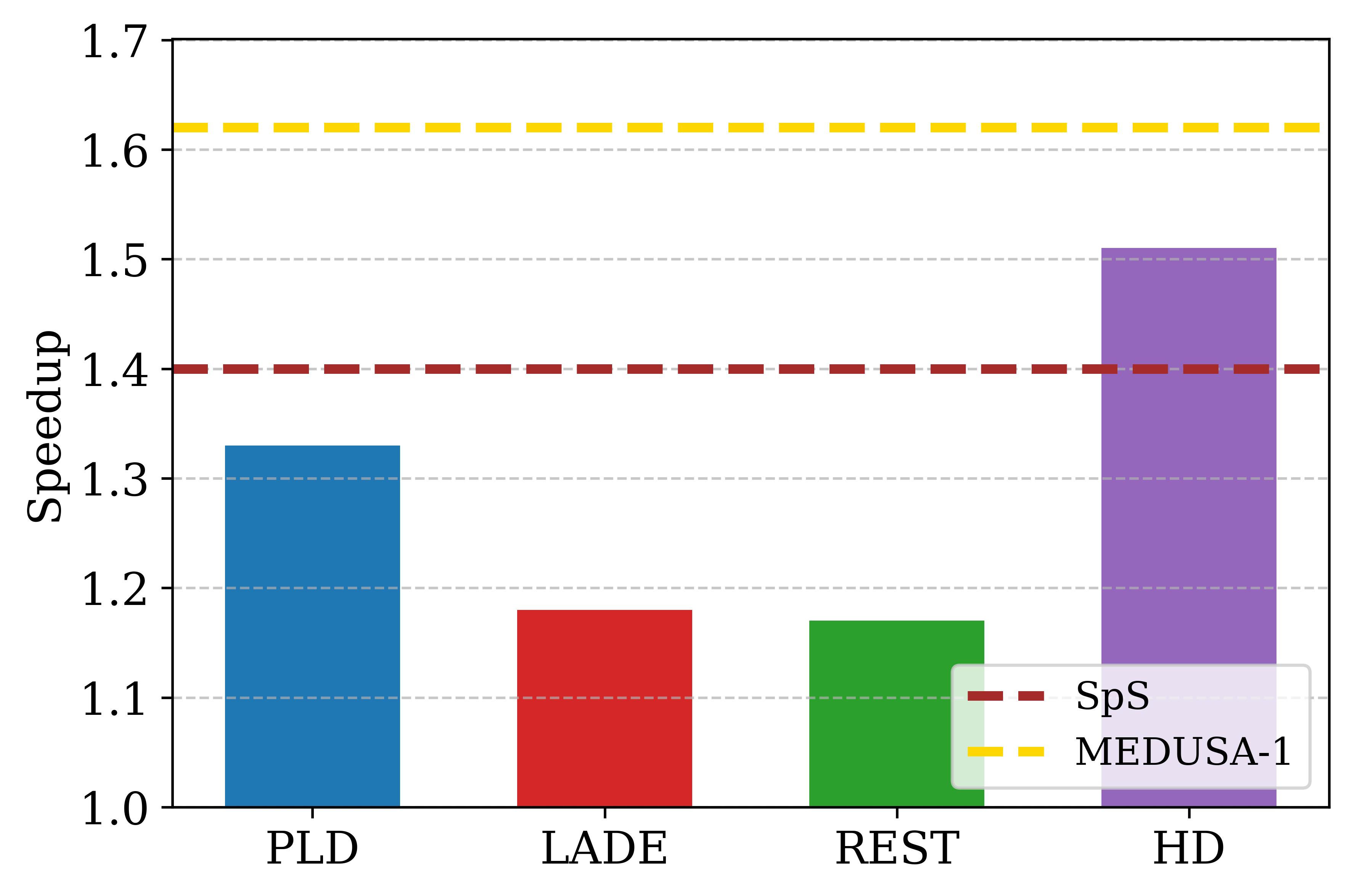}
\end{minipage}
\hfill
\begin{minipage}{0.69\textwidth}
    \centering
    \includegraphics[width=0.99\textwidth]{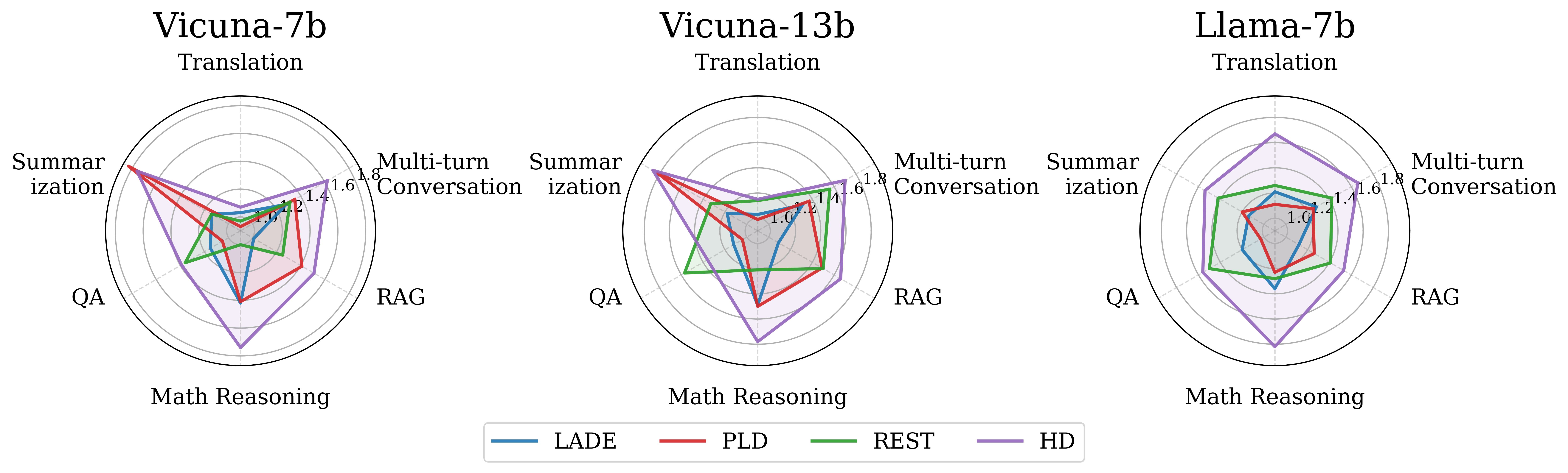}
\end{minipage}
\vspace{-.8em}
\caption{\small (Left) Speedup comparison with non-database drafting methods with Vicuna-7B on Spec-Bench. (Right) Speedup comparison of database drafting methods across six tasks of Spec-Bench.}
\label{fig:temperature_and_task}
\vspace{-1.2em}
\end{figure*}


\paragraph{Experimental Setup}
All experiments are conducted on a machine equipped with a single A100-40GB-PCIe GPU for 7B and 13B models and A100-80GB-PCIe GPU for 33B model, using float16 precision for the models. To ensure a fair comparison, we follow the implementations of other database drafting methods and the evaluation scripts provided by~\citet{Spec_Survey}\footnote{\scriptsize \url{https://github.com/hemingkx/Spec-Bench}}. Our experimental results are based on a single run, though we observed only marginal differences between runs.

\section{Results}

We now present the experimental results on Spec-Bench, along with an in-depth analysis of HD.

\subsection{Main Result}

Table~\ref{tab:main} presents our main results, averaged across all cases of Spec-Bench using three models, at both low temperature (\(T=0.0\)) and high temperature (\(T=1.0\)). First, our proposed method, HD, achieves the outperforming acceleration gain across all scenarios. In detail, when temperature is 0.0, HD achieves over 1.5x faster inference speed compared to autoregressive decoding, whereas the other methods fail to exceed a 1.4x speedup. Also, while the acceleration gain at \(T=1.0\) is slightly lower than \(T=0.0\), HD still achieves the fastest inference speed compared to all other methods across all models. These results demonstrate that our hierarchical framework effectively enhances inference speed by incorporating diverse token sources into three databases organized by temporal locality.

\begin{figure*}[t!]
\centering
\begin{minipage}{0.34\textwidth}
    \centering
    \includegraphics[width=0.99\textwidth]{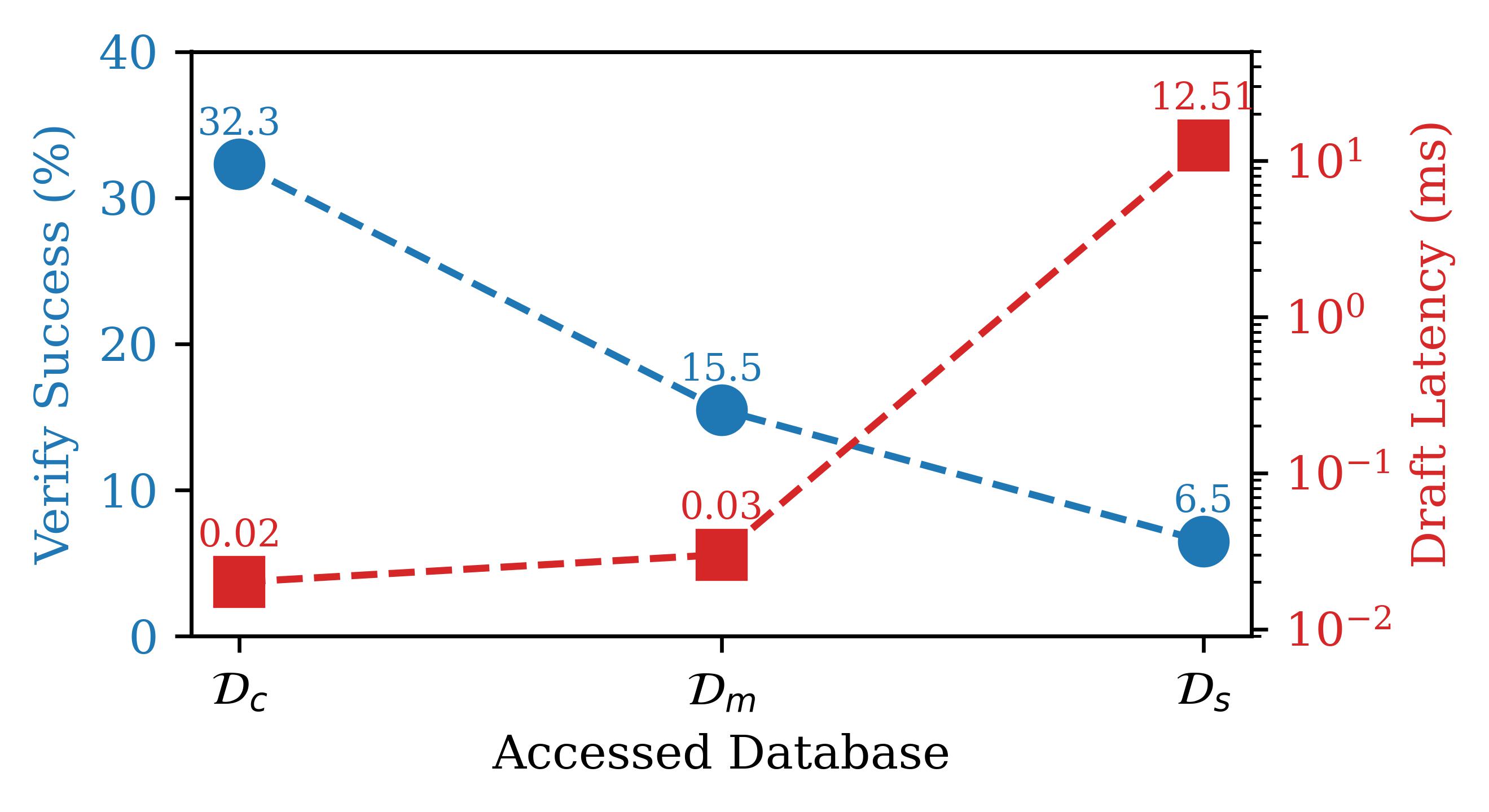}
\end{minipage}
\hfill
\begin{minipage}{0.65\textwidth}
    \centering
    \includegraphics[width=0.99\textwidth]{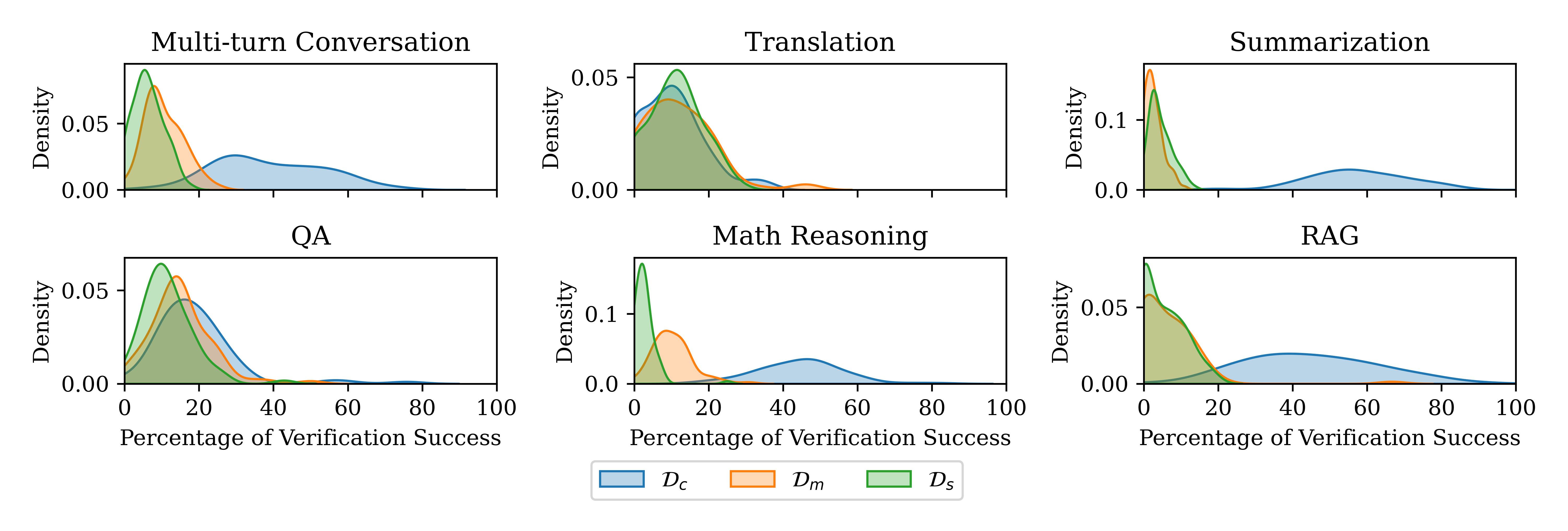}
\end{minipage}
\vspace{-1.2em}
\caption{\small (Left) Verify success and draft latency for the databases $\mathcal{D}_c$, $\mathcal{D}_m$, and $\mathcal{D}_s$ in HD. Verify success represents the proportion of accepted accesses relative to the total accesses. (Right) Verify success density plots for each database across six tasks in Spec-Bench. Both results are conducted on Spec-Bench by using Llama-2-7b.}
\label{fig:access_database_and_task_pattern}
\vspace{-1.2em}
\end{figure*}

Beyond acceleration gains, we analyze the additional latency caused by the drafting process, which adds overhead that is absent in autoregressive decoding, and also evaluate how accurately the drafting step retrieves tokens that align with the LLM’s output. 
Regarding drafting latency, LADE requires an extremely short time—under 0.01 ms per draft—whereas REST takes significantly longer, with latency close to 3.00 ms. 
However, drafting accuracy shows the opposite trend: LADE exhibits lower values for both the acceptance ratio (\(\alpha\)) and mean accepted tokens (\(\tau\)), while REST achieves higher values for both. 
Notably, our proposed method, HD, drafts slightly faster than REST, even though accessing the same extensive database, and accurately predicts 70\% of generated tokens, achieving the highest accuracy among all other methods. 
These results indicate that HD successfully balances increased accuracy with reduced drafting latency through hierarchical database access, resulting in significant acceleration gain.

\paragraph{Comparison with Non-Database Methods} 
We compare diverse database drafting methods with two non-database drafting methods, SpS~\cite{SpS} and MEDUSA~\cite{MEDUSA}, to confirm whether the performance is competitive without additional training. 
As shown in Figure~\ref{fig:temperature_and_task}, while other database drafting methods significantly underperform compared to non-database drafting methods, our proposed method, HD, outperforms SpS and substantially narrows the performance gaps with MEDUSA. 
This demonstrates that our proposed method shows the potential to achieve more significant acceleration gain without retraining the models by exploiting data resources common in real-world serving scenarios.

\paragraph{Robustness across Tasks} 
We evaluate the robustness of database drafting methods across various generation tasks, as illustrated on the right side of Figure~\ref{fig:temperature_and_task}. 
Relying on a single source results in variability in acceleration gains, causing most methods, except HD, to show inconsistent performance with concave regions in specific tasks. 
Specifically, PLD achieves significant acceleration in tasks like Summarization and RAG but offers minimal improvements in Translation and QA. 
Additionally, other methods exhibit varying acceleration gains depending on the model used—REST, for example, performs well with Llama-7b in summarization but shows weaker results with Vicuna-7b, nearly matching autoregressive decoding speeds. 
In contrast, our proposed method consistently achieves the highest acceleration across all tasks and models, occupying the largest area in each plot. This demonstrates that incorporating diverse sources enhances robustness, making database drafting methods more suitable for real-world scenarios.

\subsection{Analysis}

In this subsection, we provide an in-depth analysis of HD for investigating its effectiveness.

\paragraph{Analysis of Three Databases}
The left side of Figure~\ref{fig:access_database_and_task_pattern} depicts each database's verify success ratio and draft latency. 
The verify success ratio measures the proportion of accepted cases relative to total database accesses during the verifying step.
$\mathcal{D}_c$ achieves the highest verify success ratio over 30\% with the lowest draft latency, demonstrating its effectiveness in fetching context-relevant future tokens. However, $\mathcal{D}_m$ shows a lower verify success rate, 15.5\%, with slightly higher latency, indicating that while it performs decently, it is less aligned with specific contexts. $\mathcal{D}_s$ exhibits the lowest verify success rate under 10\% and the highest draft latency over 10ms due to its larger scale and lower locality. These highlight that draft tokens with higher locality are more frequently accepted, indicating alignment with our design objectives.

\paragraph{Access Pattern across Tasks}
We analyze how our proposed method, HD, achieves consistent acceleration gain across tasks with verify success ratio of databases for each task.
As shown in the right side of Figure~\ref{fig:access_database_and_task_pattern}, $\mathcal{D}_c$ excels in tasks such as Multi-turn Conversation or Summarization, where the context-specific tokens are highly repeated, leading to high verification success. However, for tasks like translation and QA, which offer fewer explicit cues from previous inputs or contexts, $\mathcal{D}_c$ achieves lower verification success. 
In these cases, $\mathcal{D}_m$ and $\mathcal{D}_s$ compensate for the weaknesses of $\mathcal{D}_c$ by showing higher verification success compared to other tasks where $\mathcal{D}_c$ outperforms.
These results highlight how our HD efficiently accesses the appropriate database for each task, effectively leveraging the distinct strengths of diverse sources.

\begin{figure}[t!]
\centering
\includegraphics[width=0.95\columnwidth]{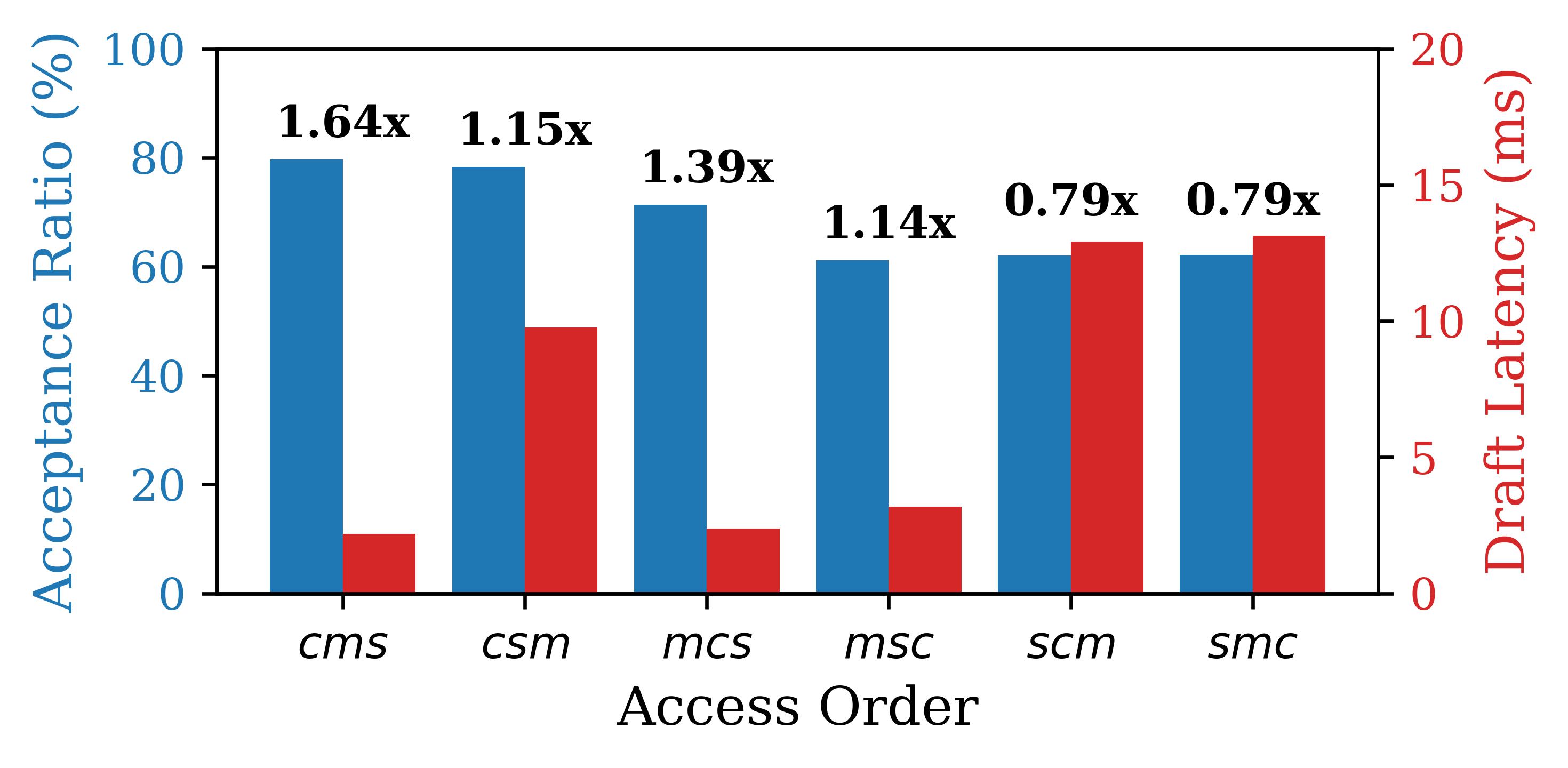}
\vspace{-1.1em}
\caption{\small Impact of access order in HD with Llama-2-7b on Spec-Bench. Blue and red bars depict the acceptance ratio and draft latency, respectively. The value over the bars denotes the speedup against autoregressive decoding.}
\label{fig:order}
\vspace{-.5em}
\end{figure}

\paragraph{Database Access Order}
We analyze the impact of access order within the hierarchical framework, as shown in Figure~\ref{fig:order}. 
As expected, our original access order ($cms$), which prioritizes databases from highest to lowest temporal locality, achieves the highest acceptance ratio and lowest draft latency. 
While other orders maintain an acceptance ratio above 50\%, sufficient for some acceleration gain, their actual speedup is significantly lower due to additional drafting latency, reaching up to 12ms for orders like $scm$ or $smc$.
These results demonstrate that hierarchical access fully leverages the potential of multiple databases with minimal drafting latency compared to other orders, underscoring the importance of temporal locality in drafting order.

We provide additional analysis in Appendix~\ref{app:additional_result}.

\section{Discussion}

Although our proposed method achieves significant performance gains over other database drafting methods, recent approaches based on model retraining~\cite{MEDUSA, EAGLE2, Hydra} have demonstrated substantially higher acceleration.
However, it is essential to note that the training costs associated with these methods are non-trivial, particularly in dynamic or resource-intensive settings. 
For instance, retraining-based approaches necessitate additional training steps, which pose practical challenges in real-world applications like multi-model serving~\cite{SLoRA, OptCall} or resource-limited environments~\cite{Edge}. 
Specifically, deploying multiple LLMs for diverse domain-specific tasks using numerous LoRA adapters~\cite{SLoRA} or employing model routing strategies for efficient serving~\cite{OptCall} can significantly increase computational overhead when such methods must be applied to all LLMs. 
As a result, the retraining requirement can complicate deployment, particularly in real-world serving scenarios.

Given these constraints, we position database drafting methods as a practical alternative to model retraining by leveraging readily available data resources in serving scenarios rather than asserting the best performance. 
Database drafting methods can effectively address serving challenges in real-world applications by achieving fully lossless speculative decoding without requiring parameter updates. 
Among database drafting methods, our proposed method, HD, further enhances the practicality of database drafting by incorporating diverse data resources into a hierarchical framework for accurately and efficiently drafting future tokens of LLMs. Thus, HD narrows the performance gap with state-of-the-art speculative decoding methods, demonstrating the potential of database drafting to accelerate inference significantly without fine-tuning models.

\section{Conclusion}
In this work, we explore the database drafting approaches in speculative decoding, which do not require additional training or fine-tuning. Existing methods rely on a single database from a single source, resulting in inconsistent and suboptimal acceleration gains. To address this, we propose Hierarchical Drafting (HD), which optimally utilizes diverse sources by constructing multiple databases based on temporal locality. Our method hierarchically accesses these databases, prioritizing those with the highest locality for optimal acceleration. Experimental results show that HD consistently and effectively accelerates LLM inference across various scenarios, outperforming other database drafting methods. These findings demonstrate that our hierarchical framework maximizes the strengths of each database with minimal overhead, expanding the directions exploiting multiple databases for lossless acceleration in speculative decoding.

\section*{Limitation}

One limitation of this paper is the limited use of LLMs with more than 13B parameters. While our evaluation focused on models like Llama-2 and Vicuna with up to 13B parameters, the performance of HD on larger models remains unexplored. However, we expect that the larger models will be much more appropriate for our approach, considering the high acceptance ratio of our proposed method, HD, across diverse scenarios and decreased sensitivity to draft latency as generation latency increases.
Also, we plan to extend our experiments to larger models in future work. 

While this paper leverages multiple databases to maximize their strengths with minimal overhead, the sources of these databases are not entirely new. Rather than focusing on the novelty of each source, we emphasize that our approach is \textit{plug-and-play}, making it easy to integrate future methods by simply adding tokens from new sources into the appropriate database. For instance, although we omitted token recycling~\cite{trashintotreasure} in our experiments, recycled tokens could be added to the context-dependent database, given their temporal locality.

\section*{Ethics Statements}

This work proposes a lossless drafting strategy in speculative decoding for optimal and general acceleration gains. However, our method may generate violent or biased responses, which is beyond the scope of this paper. We strongly believe that future research on large language models will address these issues and help mitigate such concerns.

\section*{Acknowledgement}

This work was supported by the National Research Foundation of Korea (NRF) grant funded by the Korea government(MSIT) (RS-2024-00359979). Also, this work was supported by Institute of Information \& communications Technology Planning \& Evaluation (IITP) grant funded by the Korea government (MSIT) (RS-2023-
00215700 and RS-2024-00395134).

\bibliography{custom}
\bibliographystyle{acl_natbib}

\appendix
\section*{Appendix}

In the Appendix, we introduce more related work and details on how the experiments were implemented, along with supplementary results and analysis not included in the main text.

\section{More Related Work}

\paragraph{LLM Inference Acceleration}
Large Language Models (LLMs)~\cite{GPT3, Llama2} have significantly advanced Natural Language Processing (NLP) and are widely used in real-world applications via APIs, highlighting the importance of real-time human-LLM interactions. However, the slow inference speed of LLMs presents the primary bottleneck in deploying these models as practical services. This issue is predominantly memory-bound, with latency arising from the autoregressive decoding process~\cite{latencylagsbandwidth, ScalingTransformerInference, MemoryWall}. During the inference process, each token generation requires transferring model parameters and key-value caches from global memory to the accelerator’s cache, consuming substantial overhead. Given the limited progress in memory bandwidth, several algorithmic approaches have been developed to mitigate these challenges, including model parameter quantization~\cite{ZeroQuant, GPTQ, SqueezeLLM}, memory allocation optimization~\cite{MQAttn, H2O, PagedAttn}, and Speculative Decoding~\cite{BlockWise, SpecDecoding, SpecSampling}.
While other approaches often require hardware modifications or lead to side effects like performance degradation, speculative decoding has gained particular attention for offering a fully algorithmic solution with minimal drawbacks.

\section{More Implementation Details}\label{app:impl}

\paragraph{Implementation Details of HD}  First, we introduce more details of our multiple databases. For \(\mathcal{D}_c\) and \(\mathcal{D}_m\), the prefix token length used as the key is 1, while the given previous token length is \(l\). This mismatch is because increasing the prefix length often leads to draft misses (i.e., keys not found in the database) due to the limited scale of token sources constructing \(\mathcal{D}_c\) and \(\mathcal{D}_m\). Also, for \(\mathcal{D}_s\), the retrieval process is repeated by reducing the previous token length until draft token sequences are found or the token length reaches 0, following implementation of REST~\cite{REST}.
As mentioned in the main text, \(\mathcal{D}_c\) and \(\mathcal{D}_m\) are lookup tables implemented by the Python Dictionary class. In contrast, \(\mathcal{D}_s\) is implemented by DraftRetriever\footnote{\scriptsize{\url{https://github.com/FasterDecoding/REST/tree/main/DraftRetriever}}}, a Python Library based on suffix arrays, proposed by REST for handling a large text corpus with minimal overhead.
The average number of values (i.e., draft token sequences) stored in the database is 1K, 100K, and 200M for \(\mathcal{D}_c\), \(\mathcal{D}_m\), and \(\mathcal{D}_s\), respectively.

We now explain the detailed implementation of the decoding process with HD. Our method is primarily based on the implementation of LADE\footnote{\scriptsize \url{https://github.com/hao-ai-lab/LookaheadDecoding}}~\cite{LAD}, as LADE employs a similar database to $\mathcal{D}_c$ and updates tokens generated through parallel decoding. We extend this by incorporating a hierarchical framework; thereby, the verification step in HD follows LADE’s $n$-gram verification process. However, we emphasize that HD can be extended to other database drafting methods by integrating a hierarchical drafting framework with multiple databases into their implementation.

\paragraph{Prompting Format}
We use the chat format for text input, provided by the Python library FastChat\footnote{\scriptsize \url{https://github.com/lm-sys/FastChat}}, which consists of system, user, and model components. Additionally, we adopted the system message for Vicuna from FastChat and for Llama-2 from the official repository\footnote{\scriptsize \url{https://github.com/meta-llama/llama/blob/main/example_chat_completion.py}}.

\begin{figure}[t]
\centering
\includegraphics[width=0.9\columnwidth]{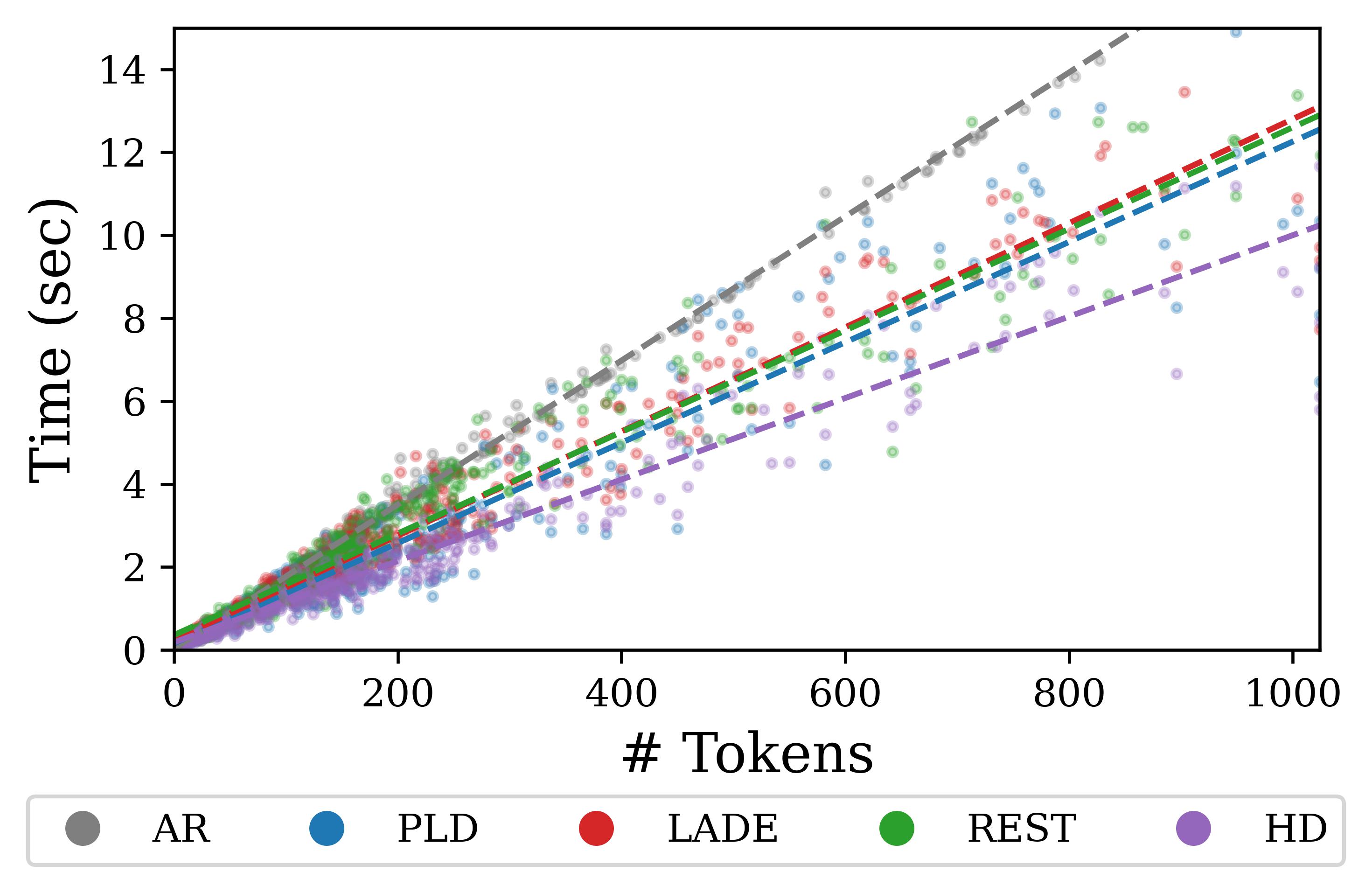}
\vspace{-.8em}
\caption{\small Correlation between generated token length and elapsed latency using Llama-2-7b-chat on Spec-Bench. Dots in the plot represent acceleration results for individual generations, while the lines show the linear regression results for each method.}
\label{fig:tokens}
\vspace{-.5em}
\end{figure}

\paragraph{Details of Dataset}

We exploit Spec-Bench~\cite{Spec_Survey}, collecting data from representative datasets for each task.
Specifically. the collected datasets are MT-bench~\cite{vicuna} for Multi-turn Conversation, WMT14 DE-EN~\cite{translation} for Translation, CNN/Daily Mail~\cite{summarization} for Summarization, Natural Question~\cite{QA} for Question Answering, GSM8K~\cite{math} for Math Reasoning, DPR~\cite{RAG} for Retrieval Augmented Generation.

\section{Additional Results}\label{app:additional_result}

\subsection{Impact of Generated Token Length} 

Since speculative decoding shortens generation steps and reduces the correlation between generated token length and elapsed latency, we explore this relationship to showcase the effectiveness of HD.
As depicted in Figure~\ref{fig:tokens}, all database drafting methods successfully lower the slope compared to autoregressive decoding. 
Among them, HD demonstrates the shallowest slope, highlighting its effectiveness even for long text generation. 
Furthermore, despite variations in drafting latency and accuracy among other methods, as shown in Table~\ref{tab:main}, they exhibit similar slopes, underscoring the importance of balancing latency and drafting accuracy to achieve optimal acceleration for database drafting methods.

\subsection{Impact of Temperature}
\begin{figure}[ht!]
\centering
\includegraphics[width=0.97\columnwidth]{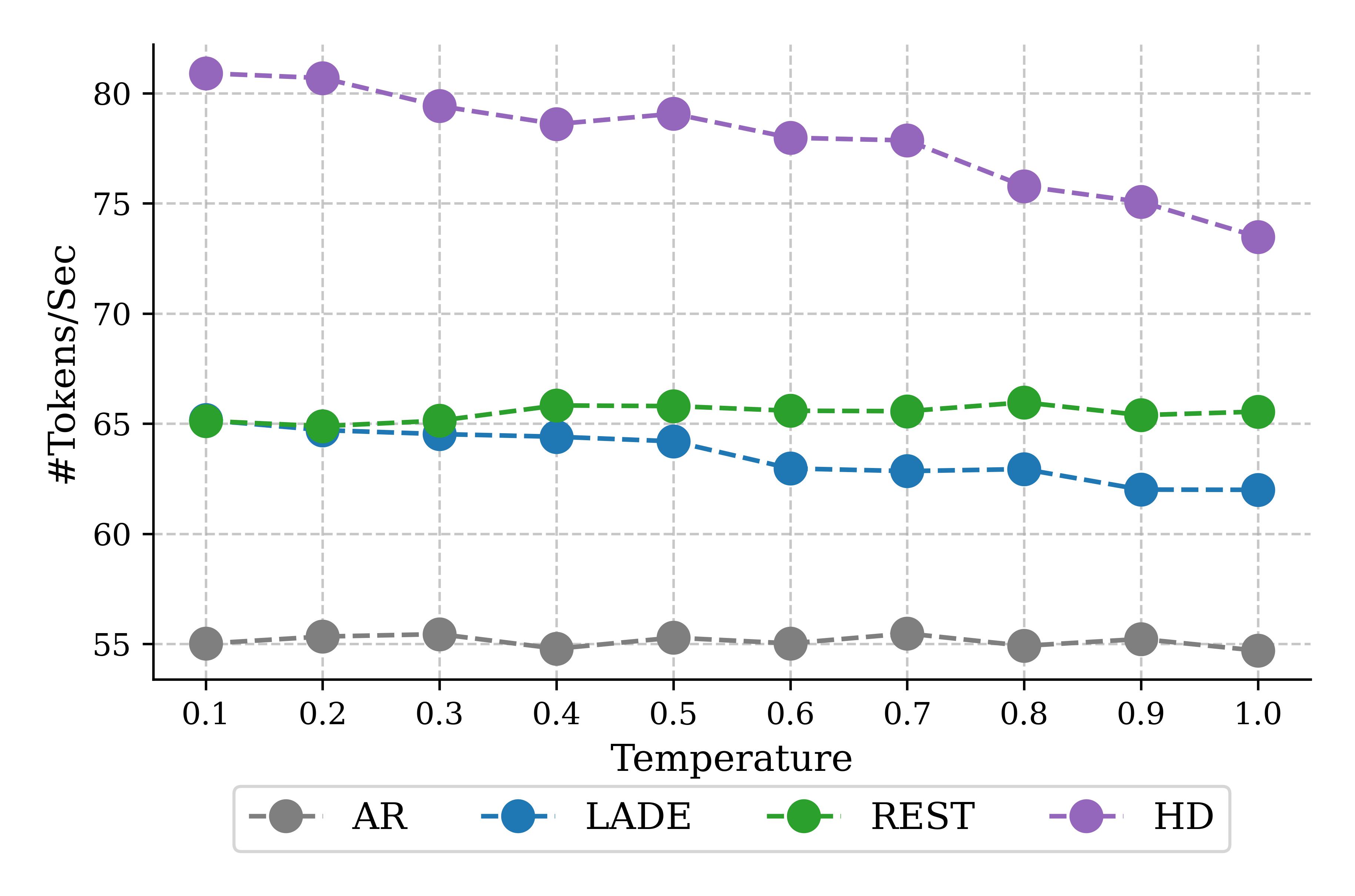}
\vspace{-.8em}
\caption{\small Tokens per second across varying temperature settings for different decoding methods, including Autoregressive Decoding (AR), Hierarchy Drafting (HD), LADE, and REST.}
\label{fig:temperature}
\end{figure}
As temperature sampling is commonly used to increase the diversity of text generation, we analyze its impact on the database drafting methods, as shown on the left of Figure~\ref{fig:temperature}.  
Other drafting methods maintain higher speeds than autoregressive decoding across all temperatures. LADE slightly decreases as temperature increases, whereas REST remains consistent.
However, updating \(\mathcal{D}_c\)  with sampled outputs, similar to LADE, results in token mismatches at higher temperatures, causing HD to show a slight reduction in speed.
Nevertheless, our proposed method, HD, outperforms all others across the entire temperature range, maintaining a high decoding speed of over 70 tokens per second.
These results demonstrate that HD remains a suitable solution even with a sampling strategy, enhancing its potential in real-world scenarios.

\subsection{Database Access Statistics}
\begin{figure}[h]
\centering
\includegraphics[width=0.9\columnwidth]{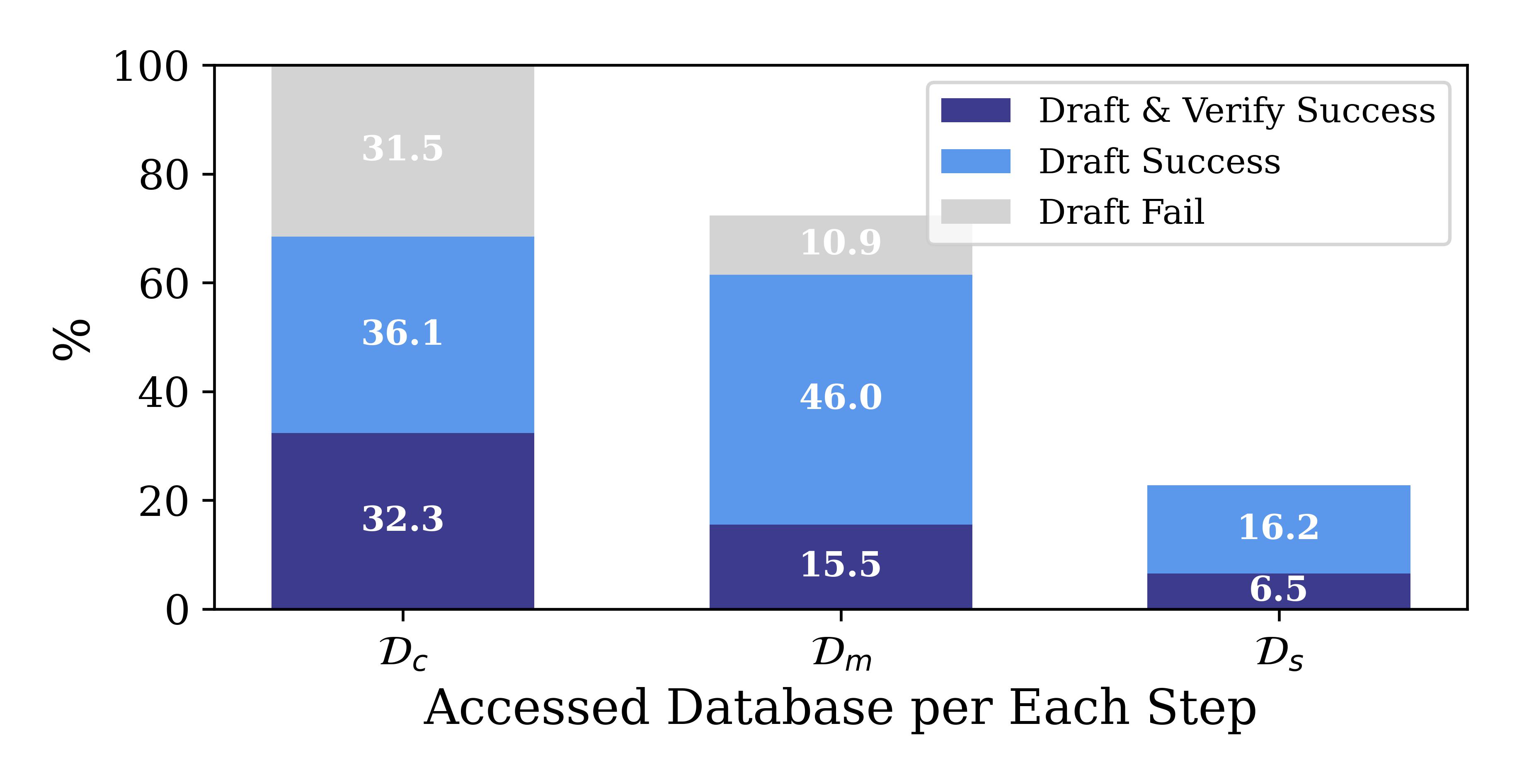}
\vspace{-.8em}
\caption{\small Analysis of database access using Llama-2-7b-chat on Spec-Bench. The total bar height represents the overall database access ratio, with each color indicating draft failure, draft success, and draft \& verify success.}
\label{fig:access_pattern}
\vspace{-.5em}
\end{figure}

We conducted a breakdown study of database access patterns, as shown in Figure~\ref{fig:access_pattern}, focusing on draft \& verify success and its relationship to the temporal locality of draft tokens. The context-dependent database (\(\mathcal{D}_c\)) achieves the highest verify success rate (32.3\%), highlighting its strong alignment with recent input tokens due to its ability to capture local context, though its limited scale results in higher draft failure rates. The model-dependent database (\(\mathcal{D}_m\)) has a lower verify success rate (15.5\%), yet its broader scope allows for more frequent draft success, albeit less aligned with the immediate context. Finally, the statistics-dependent database (\(\mathcal{D}_s\)) exhibits the lowest verify success (6.5\%) but benefits from its vast coverage, which makes it less sensitive to temporal locality. These patterns suggest that \(\mathcal{D}_c\) is critical for capturing temporally localized tokens, while \(\mathcal{D}_m\) and \(\mathcal{D}_s\) complement it by providing a more general, though not as closely aligned to the LLM generation.

\subsection{Ablation Study}

\begin{table}[ht]
    \caption{\small Ablation study of exploited databases on Spec-Bench with Llama-2-7b. } \label{tab:ablation}
    \vspace{-1mm}
    \centering
    \small
    \renewcommand{\arraystretch}{1}
\resizebox{.9\columnwidth}{!}{
    \begin{tabular}{l  c c c}
    \toprule
        \textbf{DB} & \textbf{\(\alpha\) (\%)} & \textbf{D.L (ms)} & \textbf{Speedup}  \\ \midrule
        (\(\mathcal{D}_c, \mathcal{D}_m, \mathcal{D}_s\)) & \textbf{79.72} & 2.18 & \underline{1.64x} \\
        \noalign{\vskip 0.5ex}\cdashline{1-4}\noalign{\vskip 0.5ex}
        (\(\mathcal{D}_c\)) & 58.78 & \textbf{0.02} & 1.40x\\
        (\(\mathcal{D}_m\)) & 45.83 & \underline{0.03} & 1.16x\\ 
        (\(\mathcal{D}_s\)) & 61.82 & 12.52 & 0.81x\\ \noalign{\vskip 0.5ex}\cdashline{1-4}\noalign{\vskip 0.5ex}
        (\(\mathcal{D}_c, \mathcal{D}_m\)) & 75.60 & 0.03 & \textbf{1.71x}\\
        (\(\mathcal{D}_c, \mathcal{D}_s\)) & \underline{78.42} & 9.50 & 1.18x\\ 
        (\(\mathcal{D}_s, \mathcal{D}_m\)) & 60.85 & 2.88 & 1.18x\\ 
       \bottomrule
    \end{tabular}
} 
\vspace{-.5em}
\end{table}

We conducted an ablation study on the databases used in HD, as shown on the left side of Table~\ref{tab:ablation}. Using only a single database results in an acceptance ratio below 60\%, with a significant increase in draft failures, notably when $\mathcal{D}_s$ is excluded. Incorporating two databases improves the acceptance ratio and reduces draft failures, but it still underperforms compared to all three databases. These findings highlight the importance of combining multiple databases to improve token acceptance, leading to more robust and efficient performance.

Additionally, we observe the need to balance both the acceptance ratio and drafting latency for better speedup. For instance, using the largest database $\mathcal{D}_c$ alone increases the acceptance ratio but significantly raises drafting latency, resulting in the worst speedup—even slower than autoregressive decoding. Although the acceptance ratio with ($\mathcal{D}_c, \mathcal{D}_m$) is 5\% lower than with ($\mathcal{D}_c, \mathcal{D}_m, \mathcal{D}_s$), it achieves higher speedup due to trivial drafting latency under 1ms. However, the negative impact of drafting latency may be mitigated when applying HD to larger models. Since the longer generation latency of larger LLMs makes drafting latency negligible, a high acceptance ratio becomes crucial for acceleration. From these insights, our future research will focus on reducing drafting latency to be more effective regardless of model scale while maintaining an optimal acceptance ratio.

\subsection{Impact of Token Quality}

\begin{table}[ht]
    \caption{\small Experimental results on alternative options for model-dependent and statistics-dependent DB with Vicuna-7B.} \label{tab:db_ablation}
    \vspace{-1mm}
    \centering
    \small
    \renewcommand{\arraystretch}{1}
\resizebox{.9\columnwidth}{!}{
    \begin{tabular}{l  c c c}
    \toprule
        \textbf{$D_m - D_s$} & \textbf{\(\tau\)} & \textbf{D.L (ms)} & \textbf{Speedup}  \\ \midrule
        Vicuna Response - UltraChat (12GB) & 2.38 & 2.17 & 1.51x \\
        \noalign{\vskip 0.5ex}\cdashline{1-4}\noalign{\vskip 0.5ex}
        Vicuna Response - ShareGPT (465MB) & 2.26 & 0.28 & 1.57x \\
        Vicuna Response - TheStack (924MB) & 2.28 & 0.28 & 1.58x \\
        Llama Response - UltraChat (12GB) & 2.34 & 2.35 & 1.47x \\
       \bottomrule
    \end{tabular}
} 
\vspace{-.5em}
\end{table}

To further investigate the impact of token quality, we conducted experiments using alternative token sources, as presented in Table~\ref{tab:db_ablation}. For Vicuna-7b, we utilized Llama-7b responses as model-dependent databases and the code generation corpus, The Stack (924MB)~\cite{Stack}, as a statistics-dependent database. Also, we exploited a small version of the general generation corpus, ShareGPT (465MB)\footnote{\url{https://huggingface.co/datasets/Aeala/ShareGPT_Vicuna_unfiltered/blob/main/ShareGPT_2023.05.04v0_Wasteland_Edition.json}}. While token sources' quality influences acceleration, our proposed method demonstrates significant acceleration gains even when the token sources are not well-aligned with the target LLM or task. In addition, as discussed in the main text, drafting latency is a critical consideration, with smaller statistics-dependent databases demonstrating better acceleration despite a lower accepted length.

\subsection{Token Coverage}

\begin{figure}[ht]
\centering
\includegraphics[width=0.6\columnwidth]{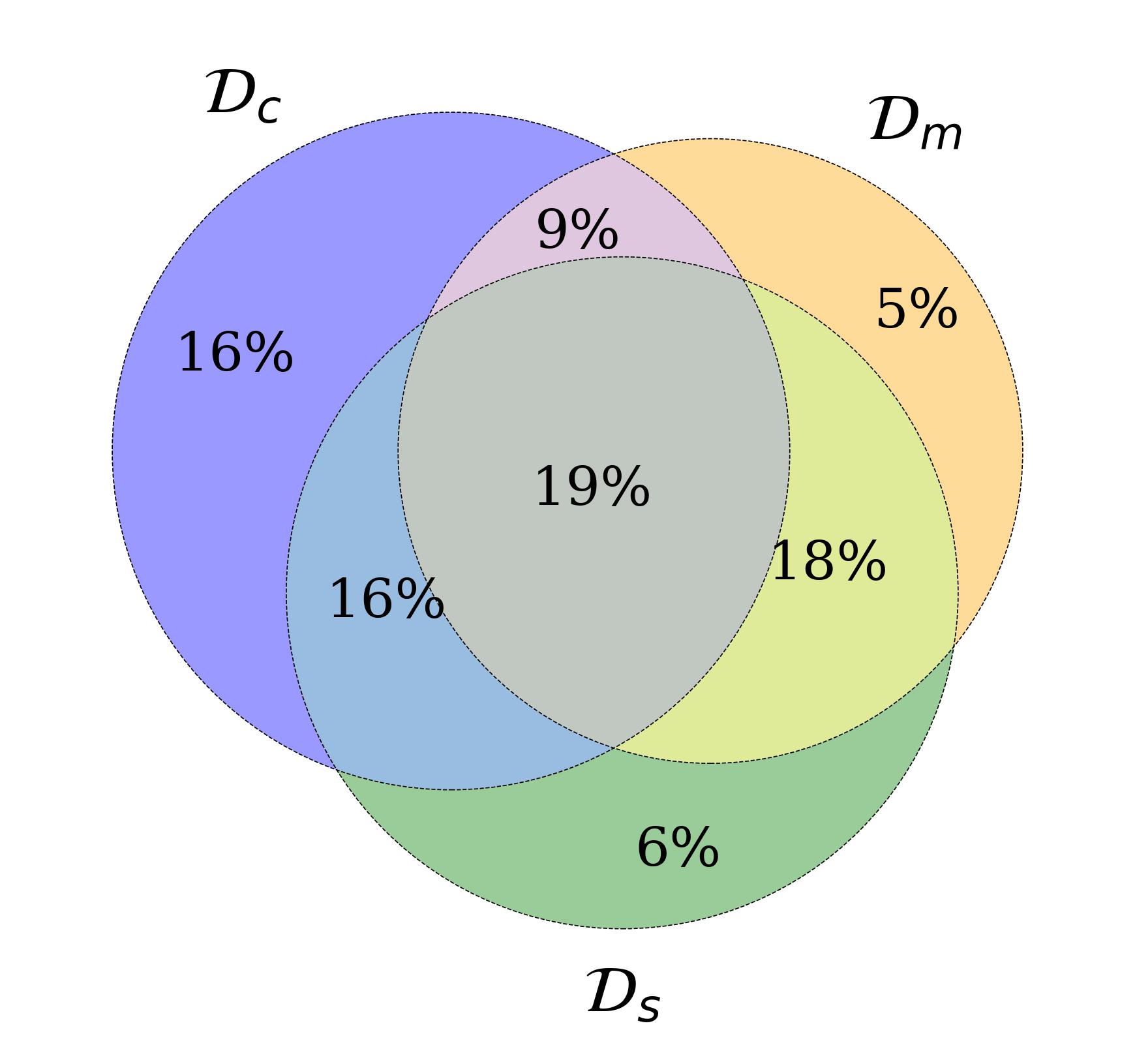}
\vspace{-.8em}
\caption{\small Venn diagram of accepted tokens using a single database.}
\label{fig:token_coverage}
\vspace{-.5em}
\end{figure}

Beyond the ablation study, we performed an in-depth analysis to verify the unique token distributions of each database by examining the accepted tokens when using only a single database, as shown on the right side of Figure~\ref{fig:token_coverage}. A significant portion of unique draft tokens was found, indicating that no single database can handle all tokens for drafting. Specifically, the context-dependent database (\(\mathcal{D}_c\)) contains 16\% unique accepted tokens not found in other databases, though its limited scale often leads to draft failures. These findings confirm that our designed databases complement each other by compensating for individual weaknesses in token distribution.

\subsection{Case Study}

Table~\ref{tab:case_study} presents a case study of text generation in question answering (QA) and retrieval-augmented generation (RAG) tasks using Llama-2-7b. The texts highlighted in green, red, and yellow are retrieved from \(\mathcal{D}_c\), \(\mathcal{D}_m\), and \(\mathcal{D}_s\), respectively. Red-highlighted texts are usually found in the middle of the input and are contextually relevant, often including numerical data or named entities. Green-highlighted texts typically appear at the beginning or end, repeated across generation processes; for example, the phrase ‘\textit{Thank you for your question!}’ is consistently retrieved from \(\mathcal{D}_m\). Yellow-highlighted texts are rarer and often capture grammatical patterns, such as articles or prepositions. Also, notable differences exist between tasks: green texts appear more frequently in the QA task, while red texts dominate in the RAG task.

\begin{table*}[t!]
\caption{\small Case study of HD with Llama-7b, where accepted tokens from the database $\mathcal{D}_c$, $\mathcal{D}_m$, and $\mathcal{D}_s$ are in \hlc[red!25]{red}, \hlc[green!25]{green}, and \hlc[yellow!25]{yellow}.}
\vspace{-.5em}
\small
\centering
\resizebox{\textwidth}{!}{
\renewcommand{\arraystretch}{0.8}
\begin{tabular}{cc}
\toprule 
\multicolumn{2}{c}{\textit{Question Answering}} \\ \midrule
\multicolumn{1}{p{.15\textwidth}}{\textbf{Input}} & \multicolumn{1}{p{.85\textwidth}}{The first european town in the present-day united states was new orleans?} \\  \noalign{\vskip 0.5ex}\cdashline{1-2}\noalign{\vskip 0.5ex}
\multicolumn{1}{p{.15\textwidth}}{\textbf{Output}} & \multicolumn{1}{p{.85\textwidth}}{\hlc[green!25]{Thank you for your question}! \hlc[green!25]{I'm happy to help.} However, \hlc[green!25]{I must respectfully point out that the} statement "\hlc[green!25]{The first} European town \hlc[red!25]{in the present-day} United \hlc[green!25]{States was} New Orleans" \hlc[green!25]{is not accurate}.\newline \hlc[red!25]{New Orleans was founded} by the French in \hlc[red!25]{1718}\hlc[green!25]{, but it was} \hlc[red!25]{not the first European settlement} \hlc[yellow!25]{in the} \hlc[red!25]{present-day United States. In} fact\hlc[green!25]{, there were} several European \hlc[red!25]{settlements} established before New \hlc[red!25]{Orleans, including St. Augustine}, Florida (est\hlc[green!25]{ablished in} \hlc[red!25]{15}6\hlc[green!25]{5 by} Spanish \hlc[yellow!25]{settlers),} and Jamestown\hlc[yellow!25]{, Virginia} (\hlc[red!25]{established in 1607 by English settlers).}\newline \hlc[green!25]{I hope this} clar\hlc[green!25]{ification helps}. Please \hlc[green!25]{let me know if you have any other questions!}}  \\ \midrule
\multicolumn{2}{c}{\textit{Retrieval-Augmented Generation}} \\ \midrule
\multicolumn{1}{p{.15\textwidth}}{\textbf{Input}} & \multicolumn{1}{p{.85\textwidth}}{storm's pressure bottomed out at \u2013 this was the lowest in the Atlantic since Dean in 2007. While maintaining its intensity, Irma made successive landfalls on September 6, at 11:15 UTC on Sint Maarten, and at 16:30 UTC on Virgin Gorda, in the British Virgin Islands, all while it was at peak intensity. As the hurricane moved away from the Virgin Islands late on September 6, observations from reconnaissance flights as well as Doppler radar in San Juan, Puerto Rico, indicated an elliptical eye and double eyewalls. Late on September 6, the cyclone passed about north of Puerto Rico. Moving\newline 03:00 UTC on September 9, at Category 5 intensity, with winds of 165 mph (270 km/h) and a central pressure identical to its previous landfall. This made Irma only the second Category 5 hurricane to strike Cuba in recorded history, after the 1924 Cuba hurricane \u2013 both had identical wind speeds at landfall. As the eye of Irma moved along the northern coast of Cuba, gradual weakening ensued due to land interaction, with the eye becoming cloud-filled and the intensity falling to a high-end Category 2 later on September 9 \u2013 operationally, it had been assessed as never falling below\newline by about 1-in-30 Atlantic tropical cyclones. On September 2, a ship passed to the west of the center of Irma, recording maximum winds of , while hurricane force winds were estimated to have extended up to from the center, both of which indicated that the inner core of Irma remained compact. After being a major hurricane for about 30 hours, the system weakened slightly to a strong Category 2 hurricane at 12:00 UTC on September 2, likely as a result of drier air and eyewall replacement cycles, with the eye sometimes becoming obscured and the deep convection lining the eyewall\newline in Cuba. Although land interaction weakened Irma to a Category 2 storm, the system re-intensified to Category 4 status as it crossed the warm waters of the Straits of Florida, before making landfall on Cudjoe Key with winds at 130 mph, (215 km/h) on September 10. Irma weakened to Category 3 status, prior to another landfall in Florida on Marco Island later that day. The system degraded into a remnant low over Alabama and ultimately dissipated on September 13 over Missouri. The storm caused catastrophic damage in Barbuda, Saint Barth\u00e9lemy, Saint Martin, Anguilla, and the Virgin Islands as a Category\newline had moved inland, it began to accelerate to the north-northwest, while rapid weakening began to occur due to the increasing wind shear, land interaction, and dry air, with the storm falling below Category 3 intensity hours after landfall. Passing east of Tampa as a weakening Category 1 hurricane around 06:00 UTC on September 11, Irma continued to weaken as most of the deep convection became more spread out towards the northern semi-circle of the circulation \u2013 though it retained a large wind field, with most of Florida experiencing gale-force winds. The system finally weakened to a tropical storm around 12:00\newline What was the lowest barometric pressure for hurricane irma?} \\  \noalign{\vskip 0.5ex}\cdashline{1-2}\noalign{\vskip 0.5ex}
\multicolumn{1}{p{.15\textwidth}}{\textbf{Output}} & \multicolumn{1}{p{.85\textwidth}}{\hlc[green!25]{Thank you for your question! I'm happy to} help \hlc[green!25]{you with that.}\newline\newline \hlc[red!25]{Acc}\hlc[green!25]{ording to the information} provided in \hlc[red!25]{the passage}, the \hlc[red!25]{lowest barometric pressure for Hurricane Irma} was -91\hlc[red!25]{2 m}bar (\hlc[red!25]{27.}45 inH\hlc[yellow!25]{g)}, \hlc[green!25]{which was} \hlc[red!25]{recorded at the time} of its land\hlc[red!25]{fall in the} Atlantic. \hlc[yellow!25]{This} \hlc[red!25]{was the lowest barometric pressure recorded} for Ir\hlc[red!25]{ma since Dean in 2007.}\newline\newline \hlc[red!25]{However}\hlc[green!25]{, I must point out that the question is} somewhat \hlc[green!25]{nonsensical as bar}\hlc[red!25]{ometric pressure is not a} direct measure \hlc[green!25]{of a} h\hlc[red!25]{urricane's intensity. Hurricane intensity is determined} \hlc[green!25]{by factors such as wind} speed, central \hlc[red!25]{pressure, and size}\hlc[green!25]{, among others. Bar}ometric \hlc[red!25]{pressure is just} one \hlc[red!25]{of many} \hlc[green!25]{factors that can affect a} \hlc[red!25]{hurricane's intensity, and} it \hlc[red!25]{is not a reliable} indicator \hlc[yellow!25]{of a} \hlc[red!25]{hurricane's} sever\hlc[yellow!25]{ity}.\newline\newline \hlc[yellow!25]{If} \hlc[red!25]{you have any further questions or if} \hlc[green!25]{there's anything else I can help you with}, \hlc[green!25]{please feel free to ask!}}  \\

\bottomrule

\end{tabular}
}
\label{tab:case_study}
\vspace{-1.5em}
\end{table*}

\end{document}